\documentclass[10pt,conference]{IEEEtran}
\IEEEoverridecommandlockouts
\usepackage{cite}
\usepackage{amsmath,amssymb,amsfonts}
\usepackage{algorithmic}
\usepackage{booktabs} 
\usepackage{graphicx}
\usepackage{caption}
\usepackage{textcomp}
\usepackage{xcolor}
\usepackage{url}
\usepackage{subcaption}

\def\BibTeX{{\rm B\kern-.05em{\sc i\kern-.025em b}\kern-.08em
    T\kern-.1667em\lower.7ex\hbox{E}\kern-.125emX}}
\begin{document}

\captionsetup{font=small, labelfont=bf}

\title{FSDETR: Frequency-Spatial Feature Enhancement for Small Object Detection
\thanks{Tao Yan* is the corresponding author.}
} 

\author{
    \IEEEauthorblockN{Jianchao Huang, Fengming Zhang, Haibo Zhu and Tao Yan*}
    \IEEEauthorblockA{\textit{School of Artificial Intelligence and Computer Science, Jiangnan University, Wuxi, China}}
    \IEEEauthorblockA{\{6233112018, 6233110051, 6233152012\}@stu.jiangnan.edu.cn, yantao.ustc@gmail.com}
}

\maketitle

\begin{abstract}
Small object detection remains a significant challenge due to feature degradation from downsampling, mutual occlusion in dense clusters, and complex background interference. To address these issues, this paper proposes FSDETR, a frequency–spatial feature enhancement framework built upon the RT-DETR baseline. By establishing a collaborative modeling mechanism, the method effectively leverages complementary structural information. Specifically, a Spatial Hierarchical Attention Block (SHAB) captures both local details and global dependencies to strengthen semantic representation. Furthermore, to mitigate occlusion in dense scenes, the Deformable Attention-based Intra-scale Feature Interaction (DA-AIFI) focuses on informative regions via dynamic sampling. Finally, the Frequency-Spatial Feature Pyramid Network (FSFPN) integrates frequency filtering with spatial edge extraction via the Cross-domain Frequency-Spatial Block (CFSB) to preserve fine-grained details. Experimental results show that with only 14.7M parameters, FSDETR achieves 13.9\% $AP_S$ on VisDrone 2019 and 48.95\% $AP_{50}^{tiny}$ on TinyPerson, showing strong performance on small-object benchmarks. The code and models are
available at https://github.com/YT3DVision/FSDETR.
\end{abstract}

\begin{IEEEkeywords}
Small object detection, Feature pyramid, Frequency–spatial feature enhancement, RT-DETR
\end{IEEEkeywords}

\vspace{-2pt} 
\section{Introduction}
\begin{figure}[t]
    \centering  
    \includegraphics[width=0.95\columnwidth]{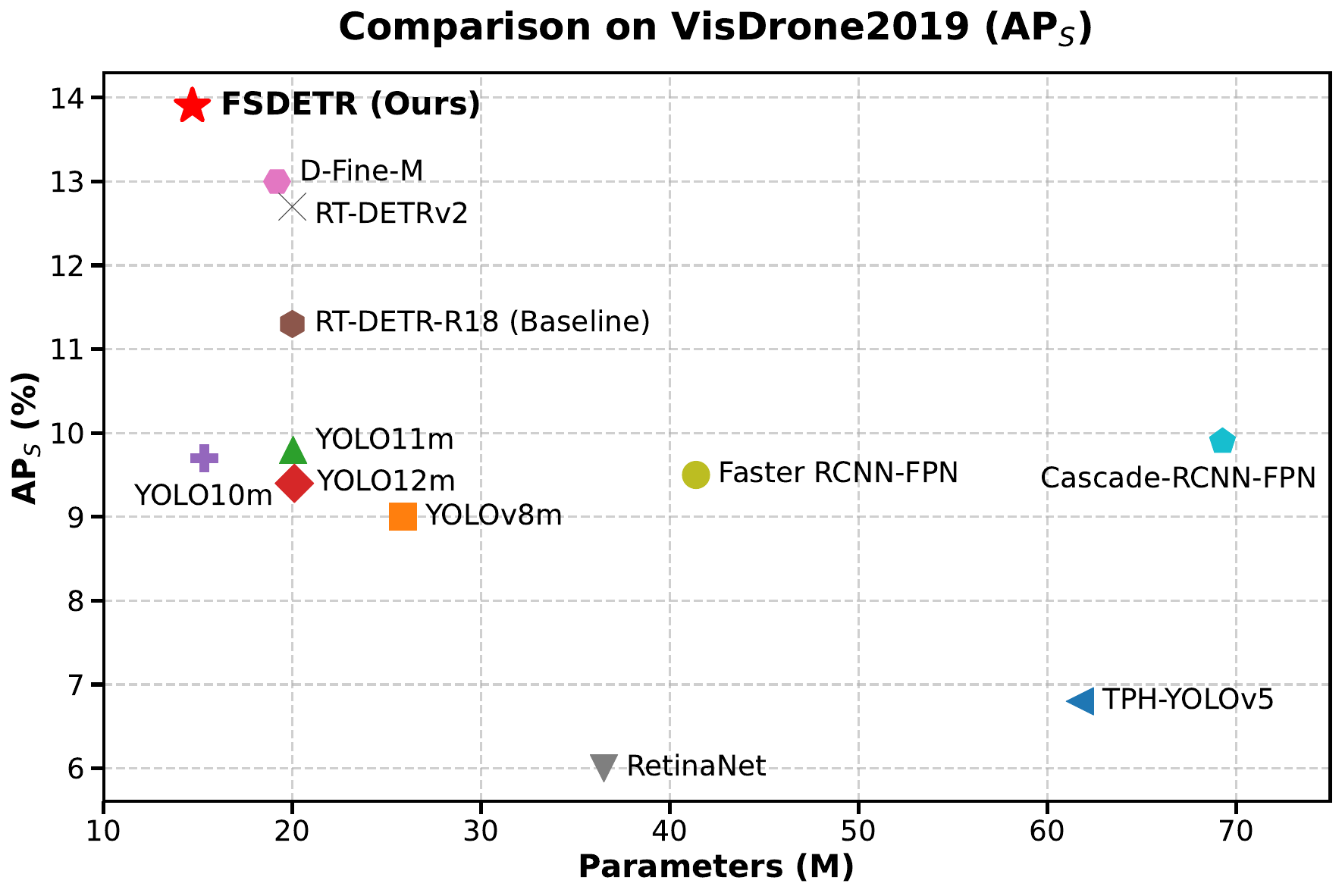}
    
    \caption{Comparison of different object detection models on VisDrone2019. FSDETR achieves the highest AP$_s$ among models with comparable parameter sizes.}
    \label{fig:scatter_comparison} 

    \vspace{-8pt}
\end{figure}

Small object detection, particularly for targets smaller than $32 \times 32$ pixels in UAV imagery, remains a challenging task due to structural information loss during deep downsampling, mutual occlusion in dense scenes, and interference from complex backgrounds. These factors severely weaken the discriminative representation of tiny targets and make accurate localization more difficult. As shown in Fig.~\ref{fig:scatter_comparison}, improving small-object detection while maintaining model efficiency is still a challenging problem in practical applications.

\begin{figure*}[t]
    \centering  
    \includegraphics[width=0.95\textwidth]{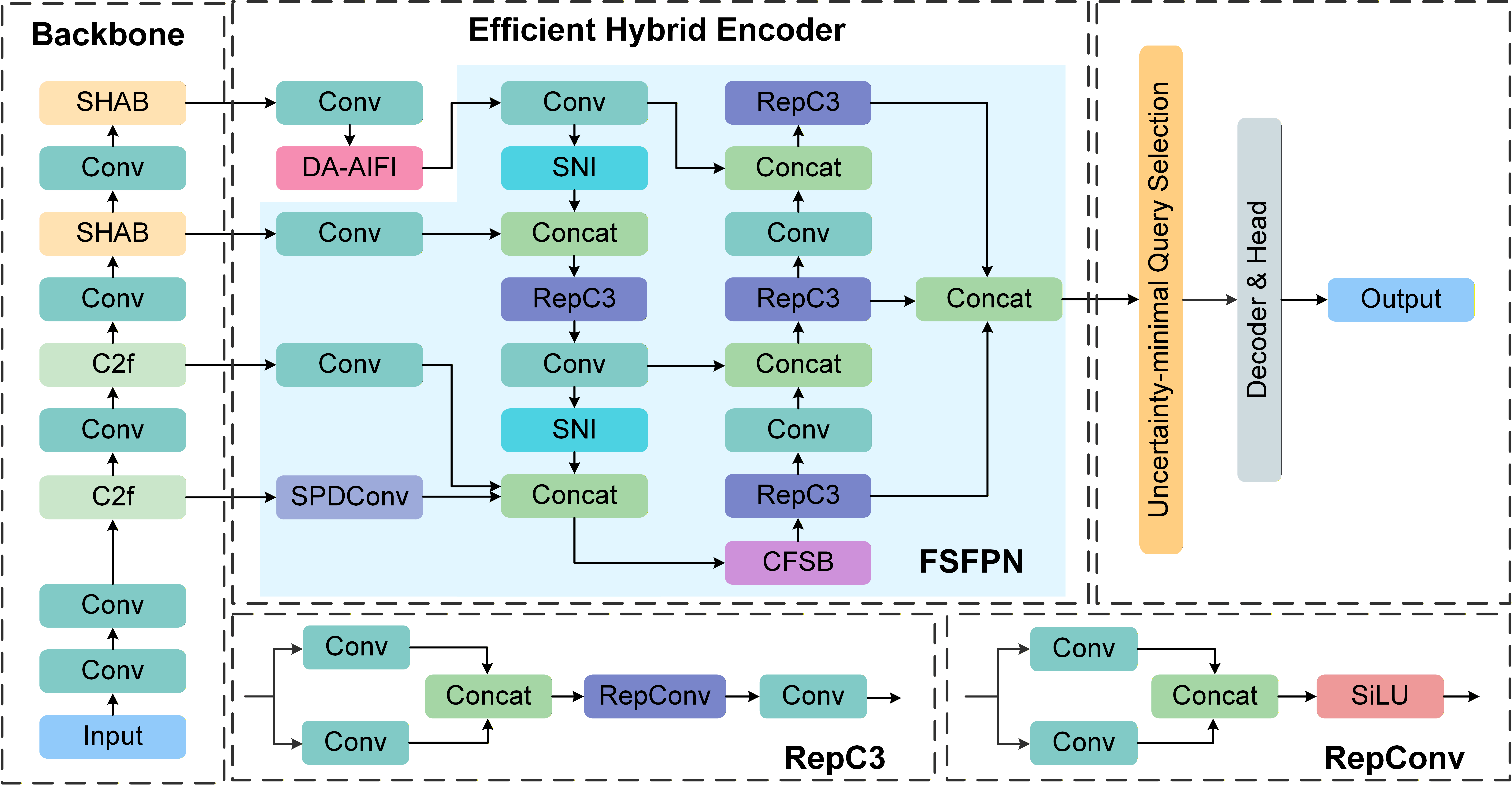}
    
    \caption{The overall architecture of FSDETR. The framework comprises a hierarchical Backbone with SHAB and C2f blocks, followed by an Efficient Hybrid Encoder. The encoder employs DA-AIFI for intra-scale interaction and FSFPN with SPDConv and RepC3 blocks for cross-scale fusion. Uncertainty-minimal Query Selection initializes object queries for the Decoder \& Head. Bottom panels illustrate detailed re-parameterized blocks (RepC3 and RepConv).}
    \label{fig:architecture} 
    \vspace{-6pt}
\end{figure*}

To address these limitations, this paper proposes FSDETR, a frequency-spatial feature enhancement framework built upon RT-DETR~\cite{zhao2023detrs}. As illustrated in Fig.~\ref{fig:architecture}, the framework introduces three targeted improvements. First, the Spatial Hierarchical Attention Block (SHAB) is integrated into the backbone to strengthen long-range structural modeling. Second, the Deformable Attention-based Intra-scale Feature Interaction (DA-AIFI) module is employed to enhance geometric alignment for sparse targets. Third, the Frequency-Spatial Feature Pyramid Network (FSFPN) and its core unit, the Cross-domain Frequency-Spatial Block (CFSB), are designed to couple spatial edge extraction with learnable frequency filtering, thereby improving fine-grained detail preservation.

Extensive evaluations on the VisDrone 2019~\cite{Du_2019_ICCV} and TinyPerson~\cite{yu2020scale} datasets demonstrate that FSDETR achieves 13.9\% $AP_S$ and 48.95\% $AP_{50}^{tiny}$, showing strong performance on small-object benchmarks. These results validate the effectiveness of synergistic frequency-spatial modeling for small object detection. The main contributions of this paper are summarized as follows:

\begin{itemize}
    \item We design FSFPN and CFSB to enhance multi-scale aggregation and frequency-spatial feature representation for small objects.
    
    \item We integrate SHAB and DA-AIFI to improve structural modeling and adaptive feature interaction with limited computational overhead.
    
    \item Extensive experiments on the VisDrone 2019 and TinyPerson datasets demonstrate that FSDETR achieves competitive performance with clear gains on key small-object detection metrics.
\end{itemize}

\section{Related Work}

Small object detection remains challenging because tiny targets usually contain limited discriminative pixels and are easily affected by background clutter, occlusion, and resolution degradation. Existing studies mainly improve performance through two directions. One direction enhances detail preservation by using multi-scale feature fusion or modified downsampling operations, such as path aggregation strategies and SPD-based convolutional designs~\cite{liu2018path, spd-conv2022, Liu2025GCGP}. The other direction introduces task-specific designs for dense or tiny targets to improve localization and feature discrimination under complex scenes~\cite{Zhu_2021_ICCV, xu2022rfla}. Although these methods have shown effectiveness, preserving fine-grained details during deep feature abstraction remains a critical difficulty.

Transformer-based detectors have offered a new paradigm for object detection by modeling long-range dependencies and reducing hand-crafted components. RT-DETR~\cite{zhao2023detrs} establishes a DETR-style baseline for real-time detection, and subsequent studies such as RT-DETRv2~\cite{lv2024rtdetrv2improvedbaselinebagoffreebies}, CCD-DETR~\cite{Wang2025CCD}, and FDSI-RTDETR~\cite{Guo2025FDSI} further improve feature interaction and representation capability through sparse sampling and lightweight designs. Meanwhile, frequency-domain learning has shown advantages in capturing patterns and preserving high-frequency information. Representative studies such as FFC~\cite{chi2020fast} and GFNet~\cite{rao2021global} show that frequency-based operators can complement spatial representations by enhancing texture and structural cues. However, most DETR-based and frequency-domain methods still mainly rely on spatial-domain detection pipelines or are designed for general visual representation learning, and their integration into small object detectors remains insufficiently explored. This gap motivates the proposed FSDETR, which introduces frequency-spatial collaborative modeling into a DETR-based framework.

\section{METHOD}
\subsection{Overall Architecture}

FSDETR follows the end-to-end detection paradigm of RT-DETR~\cite{zhao2023detrs} while introducing targeted improvements in feature extraction and fusion for small object detection. As shown in Fig.~\ref{fig:architecture}, the backbone adopts an improved CSPNet~\cite{wang2020cspnet} with sequential convolutions and C2f modules~\cite{wang2024yolov10} to produce multi-scale features $\{P_2, P_3, P_4, P_5\}$. SHAB is inserted into deep stages to enhance long-range structural modeling with limited overhead.

The efficient hybrid encoder consists of DA-AIFI and FSFPN. DA-AIFI applies deformable attention~\cite{zhu2021deformable} on high-level features to improve spatial alignment by adaptive sparse sampling, while FSFPN employs CFSB to enhance high-frequency cues such as edges and textures~\cite{chi2020fast}. The enhanced features are then fed into the standard query selection module and Transformer decoder for prediction~\cite{zhao2023detrs, zhang2022dino}.

\subsection{Spatial Hierarchical Attention Block}
\begin{figure}[htbp]
  \centering
  \includegraphics[width=0.8\linewidth]{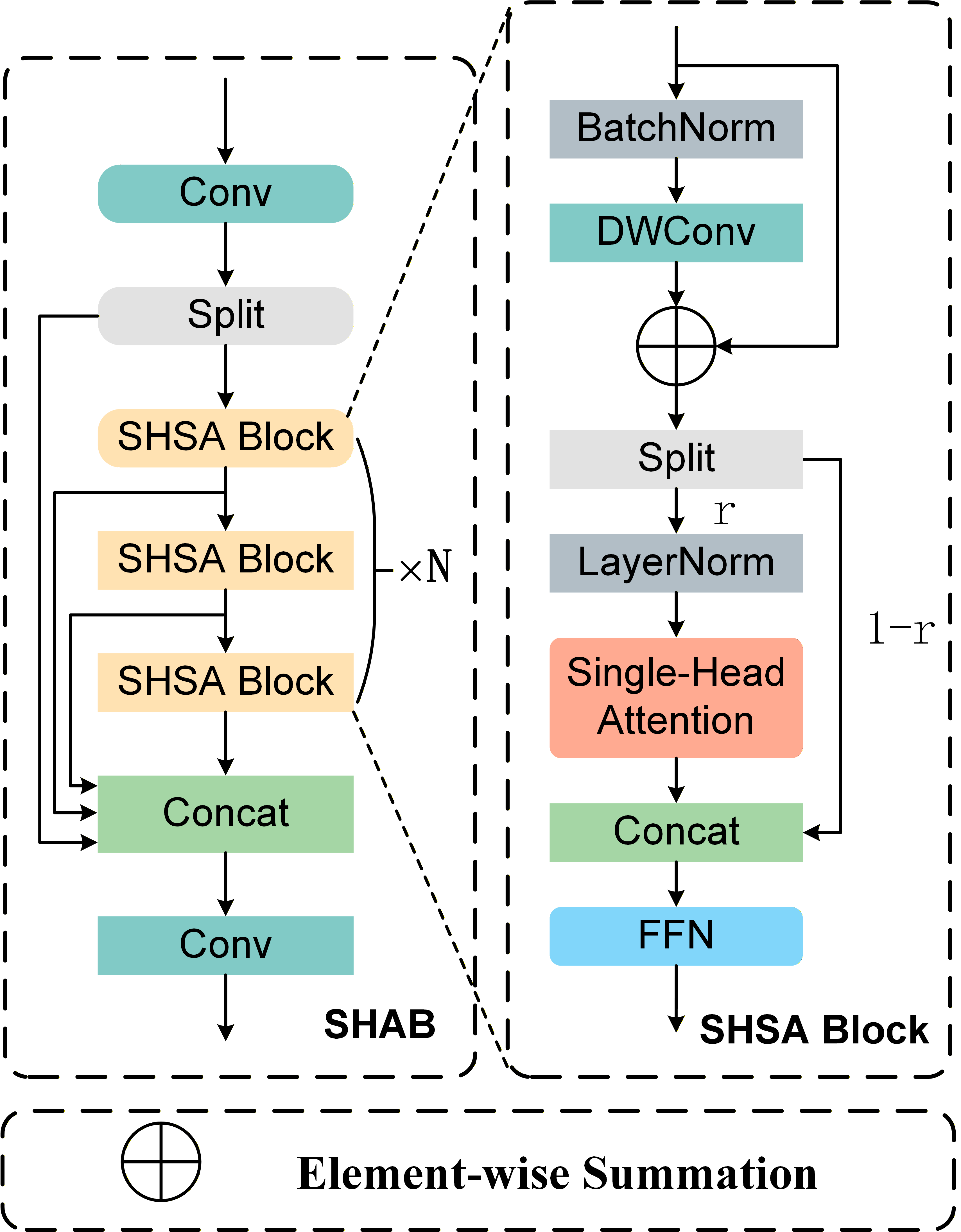}
  \caption{The structure of Spatial Hierarchical Attention Block.}
  \label{fig:shab}

  \vspace{-6pt}
\end{figure}

To mitigate feature erosion caused by progressive spatial downsampling~\cite{spd-conv2022} and address the computational redundancy in long-range dependency modeling~\cite{zhao2023detrs}, the improved CSPNet backbone integrates the Spatial Hierarchical Attention Block (SHAB), which is designed by refining the C2f structure~\cite{Jocher_Ultralytics_YOLO_2023}. As illustrated in Fig.~\ref{fig:shab}, SHAB inherits the efficient cross-stage partial topology of the C2f module but substitutes its conventional bottleneck units with SHSA Blocks. Specifically, this design maintains a dual-branch pathway to facilitate gradient flow, while the embedded SHSA Blocks ($N=1$) introduce a targeted self-attention mechanism to enhance deep representations with global context.

The core design of this block is a symmetric channel partitioning strategy with a split ratio $r=0.5$. By applying attention modeling exclusively to a subset of channels to capture global context while preserving the remaining channels via identity mapping, SHAB facilitates the establishment of long-range spatial dependencies with reduced computational complexity~\cite{chen2023run}. This structural design effectively retains fine-grained spatial details and mitigates the erosion of small-object features in deep stages without incurring significant computational overhead~\cite{wang2021pyramid}.

\subsection{Deformable Attention Intra-scale Feature Interaction}

Standard Intra-scale Feature Interaction (AIFI) modules often suffer from background noise and limited geometric flexibility due to global Multi-Head Self-Attention~\cite{zhao2023detrs, Zhu_2021_ICCV, yu2020scale}. To mitigate these constraints, we integrate DA-AIFI to adaptively focus on informative regions. Adopting the standard deformable attention formulation~\cite{zhu2021deformable}, the module aggregates features from a small set of $K$ sampling points around a reference $p_q$, rather than the entire feature map. Specifically, the output is computed as a weighted sum of sampled features, dynamically modulated by learnable offsets $\Delta p_{mqk}$ and attention weights $A_{mqk}$ across $M$ heads. By employing this sparse sampling mechanism, DA-AIFI effectively suppresses noise interference and reduces interaction complexity from $O(N^2)$ to $O(NK)$ (where $K \ll N$). This synergy of dynamic receptive fields and geometric alignment significantly enhances geometric adaptability, leading to improved small object localization and detection precision.

\begin{table*}[htbp]
\centering
\caption{Comparison results on VisDrone 2019 dataset}
\label{tab:visdrone}
\resizebox{0.9\textwidth}{!}{%
\begin{tabular}{lcccccc}
\toprule
Model & Parameters (M) & $AP_{50}$ (\%) & $AP$ (\%) & $AP_{S}$ (\%) & $AP_{M}$ (\%) & $AP_{L}$ (\%) \\
\midrule
RT-DETR-R18 (Baseline)~\cite{zhao2023detrs} & 20.0 & 36.3 & 20.8 & 11.3 & 30.5 & 37.8 \\
RetinaNet~\cite{Lin2017FocalLF} & 36.51 & 27.6 & 16.4 & 6.0 & 27.4 & 42.7 \\
Faster RCNN-FPN~\cite{lin2017feature} & 41.39 & 32.9 & 19.4 & 9.5 & 30.9 & 42.9 \\
Cascade-RCNN-FPN~\cite{cai2018cascade} & 69.29 & 32.6 & 19.7 & 9.9 & 30.9 & 40.6 \\
YOLOv8m~\cite{Jocher_Ultralytics_YOLO_2023} & 25.85 & 33.2 & 19.0 & 9.0 & 29.4 & 41.7 \\
YOLOv10m~\cite{wang2024yolov10} & 15.32 & 34.5 & 19.5 & 9.7 & 30.0 & 41.4 \\
YOLO11m~\cite{yolo11_ultralytics} & 20.04 & 35.0 & 20.3 & 9.8 & 31.2 & 41.3 \\
YOLO12m~\cite{tian2025yolo12} & 20.11 & 33.6 & 19.2 & 9.4 & 29.8 & 38.6 \\
D-Fine-M~\cite{peng2024dfine} & 19.19 & \textbf{40.7} & \textbf{23.3} & 13.0 & 33.5 & 46.4 \\
TPH-YOLOv5~\cite{Zhu_2021_ICCV} & 61.73 & 29.6 & 17.3 & 6.8 & 28.3 & \textbf{52.8} \\
RT-DETRv2~\cite{lv2024rtdetrv2improvedbaselinebagoffreebies} & 20.0 & 39.1 & 22.2 & 12.7 & 32.1 & 45.6 \\
\textbf{FSDETR (Ours)} & \textbf{14.7} & 40.5 & 22.7 & \textbf{13.9} & \textbf{33.8} & 43.5 \\
\bottomrule
\end{tabular}%
}
\end{table*}

\begin{table*}[htbp]
\centering
\caption{Comparison results on TinyPerson dataset}
\label{tab:tinyperson}
\resizebox{0.9\textwidth}{!}{%
\begin{tabular}{lcccccccc}
\toprule
Model & $AP_{50}^{tiny1}$ & $AP_{50}^{tiny2}$ & $AP_{50}^{tiny3}$ & $AP_{50}^{tiny}$ (\%) & $AP_{50}^{small}$ (\%) & $AP_{25}^{tiny}$ & $AP_{75}^{tiny}$ \\
\midrule
RT-DETR-R18 (Baseline)~\cite{zhao2023detrs} & 24.56 & 43.22 & 48.30 & 42.44 & 58.06 & 69.33 & 4.35 \\
RetinaNet~\cite{Lin2017FocalLF} & 11.47 & 36.36 & 43.32 & 30.82 & 43.38 & 57.33 & 2.64 \\
Faster RCNN-FPN~\cite{lin2017feature} & 27.21 & 48.26 & 52.48 & 43.55 & 55.69 & 63.01 & 4.75 \\
Adaptive FreeAnchor~\cite{2019FreeAnchor} & 24.92 & 48.01 & 51.23 & 41.36 & 53.36 & 63.73 & 4.00 \\
YOLOv8m~\cite{Jocher_Ultralytics_YOLO_2023} & 21.45 & 42.80 & 44.10 & 38.65 & 54.20 & 67.40 & 3.20 \\
YOLOv10m~\cite{wang2024yolov10} & 22.10 & 43.45 & 45.20 & 39.72 & 55.45 & 68.15 & 3.45 \\
YOLO11m~\cite{yolo11_ultralytics} & 22.50 & 45.10 & 46.15 & 40.58 & 56.30 & 69.90 & 3.68 \\
D-Fine-M~\cite{peng2024dfine} & 26.92 & 48.55 & 51.42 & 47.28 & \textbf{63.10} & 70.55 & 4.95 \\
RFLA~\cite{xu2022rfla} & 29.70 & 49.35 & 55.34 & 45.31 & 57.53 & 65.07 & 4.97 \\
\textbf{FSDETR (Ours)} & \textbf{31.85} & \textbf{50.12} & \textbf{56.45} & \textbf{48.95} & 62.58 & \textbf{72.80} & \textbf{5.12} \\
\bottomrule
\end{tabular}%
}
\vspace{-6pt}
\end{table*}

\subsection{Frequency-Spatial Feature Pyramid Network}

FSFPN addresses semantic dilution and aliasing in small object detection by reconceptualizing the conventional pyramid as a signal restoration and enhancement module. Specifically, for low-level features $P_2$, SPDConv~\cite{spd-conv2022} replaces strided convolution to preserve critical spatial details via pixel remapping. In the top-down pathway, Soft Nearest Neighbor Interpolation (SNI) replaces standard sampling\cite{li2024rethinking}, utilizing $\alpha = \text{Res}(X)/\text{Res}(Y)$ to adaptively mitigate the suppression of fine-grained textures.

\begin{figure}[htbp]
  \centering
  \includegraphics[width=0.65\linewidth]{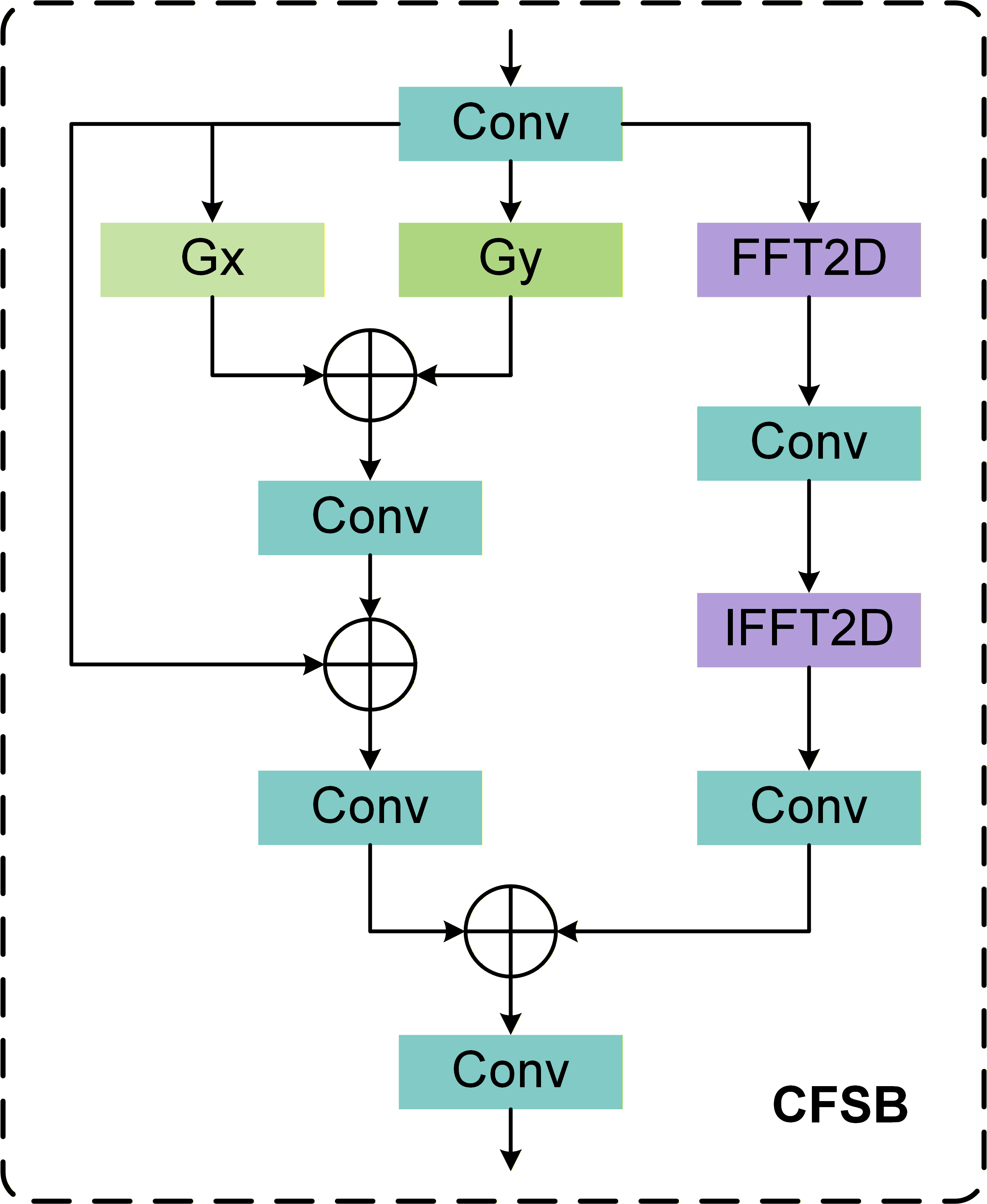}
  \caption{The structure of Cross-domain Frequency-Spatial Block.}
  \label{fig:cfsb}
\end{figure}

The core component of FSFPN is CFSB, illustrated in Fig.~\ref{fig:cfsb}. This module functions as a hybrid perception unit that synergistically processes multi-dimensional information through parallel branches~\cite{chi2020fast}. The spatial branch utilizes Scharr operators to capture local structural cues via horizontal ($G_x$) and vertical ($G_y$) components. The gradient-fused representation, denoted as $X_{grad} = G_x(X) + G_y(X)$, emphasizes object contours (or highlights geometric boundaries). where + denotes element-wise addition. To enhance feature robustness and preserve fine-grained spatial details, an internal residual connection is integrated, yielding the spatial enhanced feature:

\begin{equation}
X_{spatial} = \text{Conv}(\text{Conv}(X_{grad}) + X).
\end{equation}

The frequency-domain branch captures global spectral distributions via the 2D Discrete Fourier Transform (DFT)~\cite{rao2021global}. In this domain, high-frequency components correspond to fine textures, while low-frequency components represent macro contours. A learnable convolutional layer acts as a filter mask $\mathcal{M}$ to suppress background noise and enhance high-frequency target signals before feature reconstruction via the Inverse DFT (IDFT)~\cite{chi2020fast}. The reconstructed frequency feature $X_{freq}$ is formulated as:

\begin{equation}
X_{freq} = \text{Conv}(\mathcal{F}^{-1}(\text{Conv}(\mathcal{F}(X)))),
\end{equation}
where $\mathcal{F}(\cdot)$ and $\mathcal{F}^{-1}(\cdot)$ denote the 2D DFT and its inverse operation, respectively.

Finally, CFSB employs an adaptive fusion mechanism to integrate dual-domain information. The spatial and frequency branches are integrated through element-wise summation followed by a final convolutional layer for cross-channel interaction. This mechanism allows the model to balance locally precise edge structures with globally significant frequency patterns, yielding the final output: 
\begin{equation} X_{out} = \text{Conv}(X_{spatial} + X_{freq}). \end{equation}
By effectively modeling the interaction between spatial and frequency domains, CFSB significantly enhances the representation of small objects in complex environments.

\subsection{Optimization of Loss Function}

To enhance robustness against scale variations and sample imbalance, the total loss $\mathcal{L}$ integrates classification, regression, and geometric constraints:
\begin{equation} \mathcal{L} = \lambda_{cls} \mathcal{L}_{VFL} + \lambda_{L1} \mathcal{L}_{L1} + \lambda_{IoU} \mathcal{L}_{Focaler\text{-}EIoU},
\end{equation}
where $\lambda_{cls}$, $\lambda_{L1}$, and $\lambda_{IoU}$ denote hyperparameters balancing task contributions. For classification, Varifocal Loss (VFL)~\cite{zhang2021varifocalnet} is adopted via IoU-aware weighting to prioritize high-quality positives. For localization, $L_1$ loss is coupled with Focaler-EIoU loss~\cite{zhang2022focal} to reconstruct gradients for challenging low-overlap samples. This multi-task optimization facilitates fine-grained spatial alignment and mitigates background noise for small objects.

\begin{figure*}[htbp]
    \centering
    \includegraphics[width=0.24\textwidth]{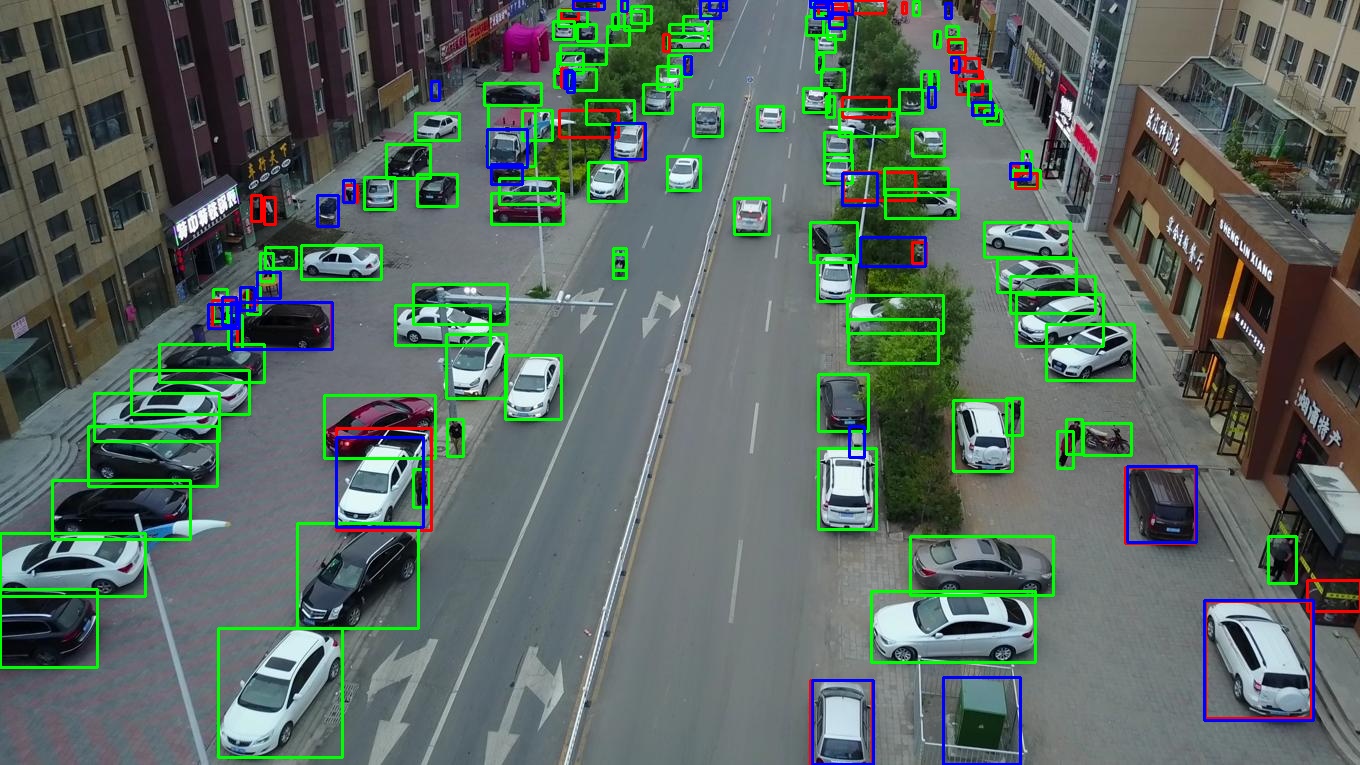}
    \includegraphics[width=0.24\textwidth]{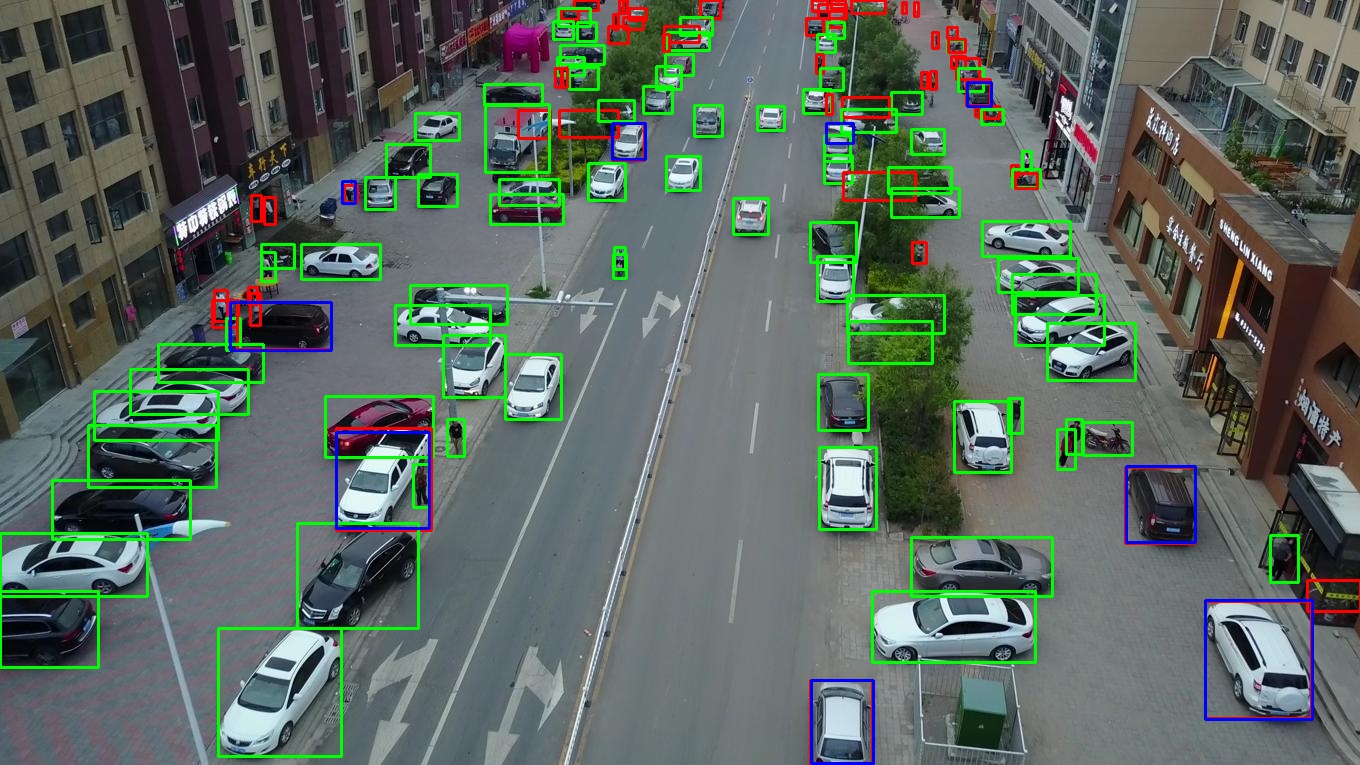}
    \includegraphics[width=0.24\textwidth]{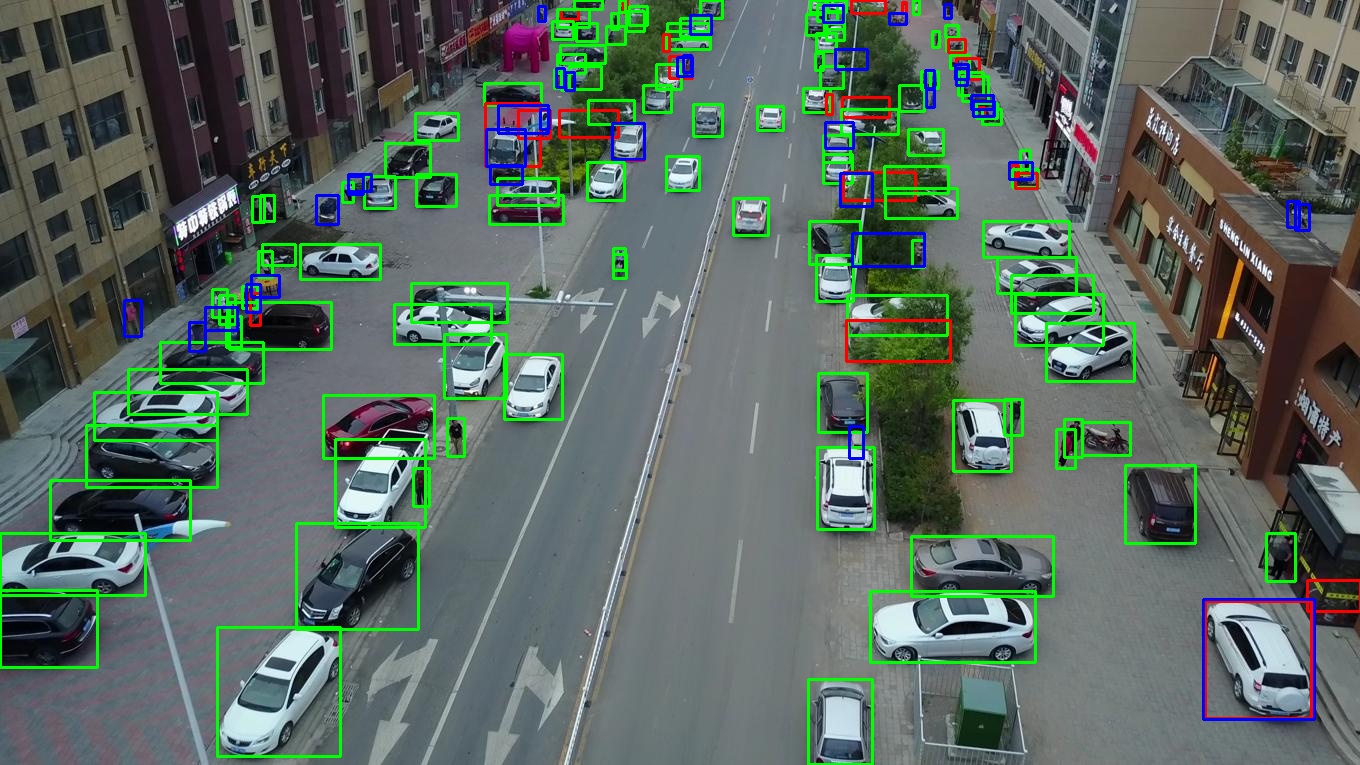}
    \includegraphics[width=0.24\textwidth]{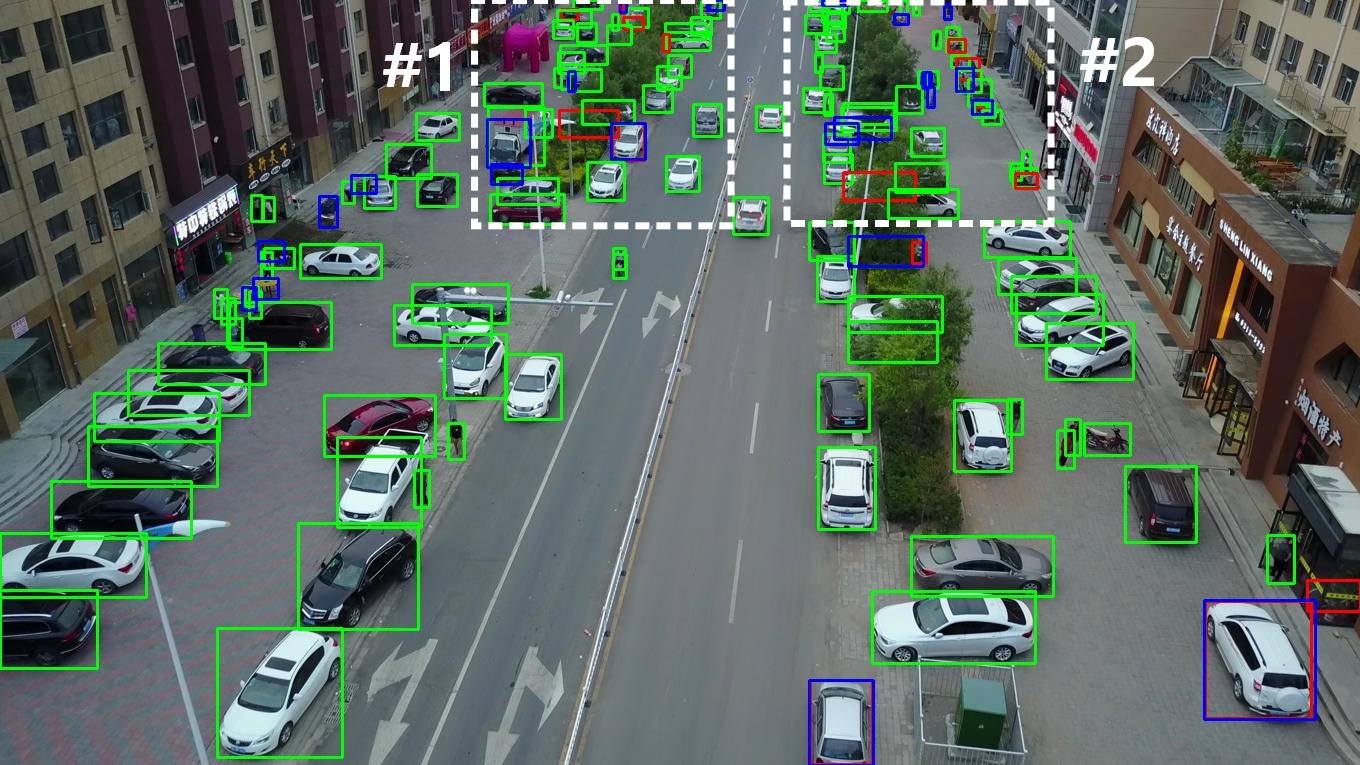}
    
    \vspace{1pt}
    
    \begin{minipage}{0.24\textwidth}
        \centering
        \includegraphics[width=\linewidth]{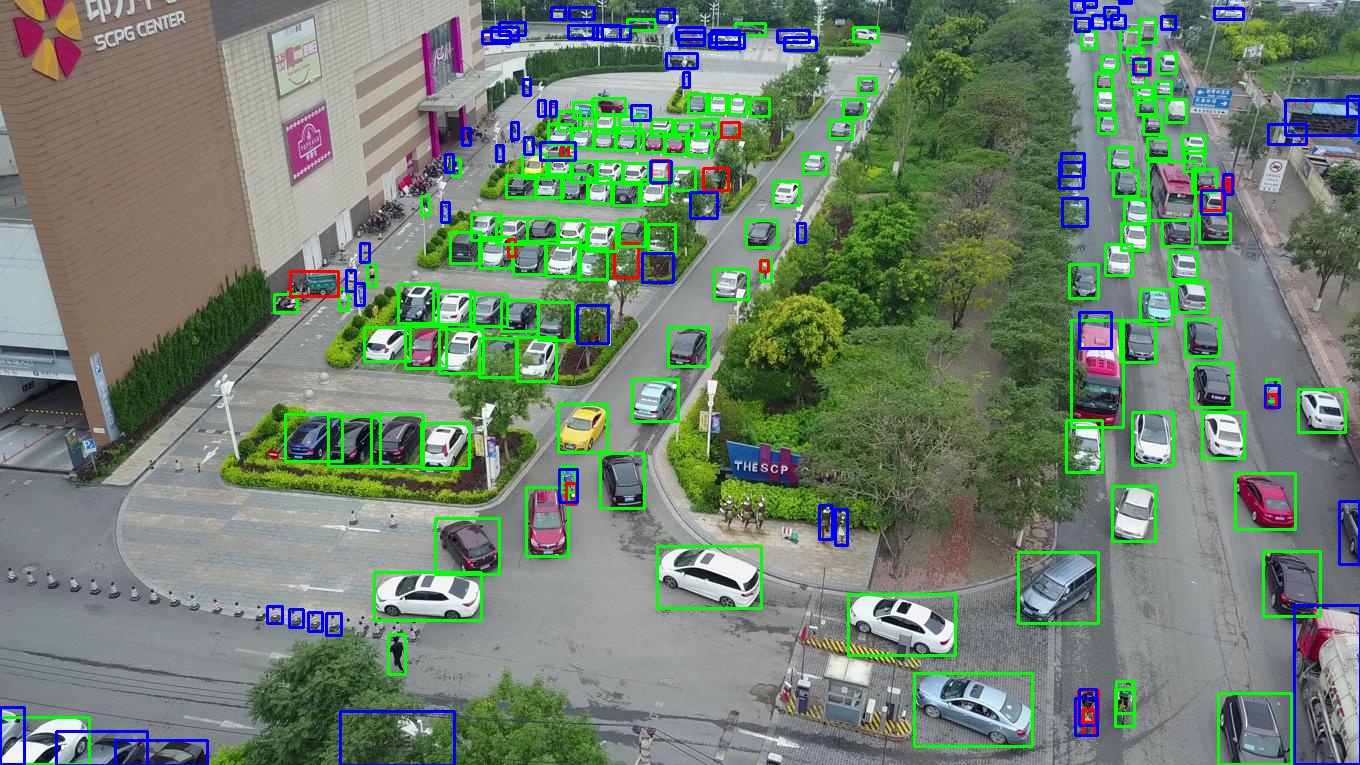}
        \vfill
        \scriptsize (a) rtdetr (Baseline)
    \end{minipage}
    \begin{minipage}{0.24\textwidth}
        \centering
        \includegraphics[width=\linewidth]{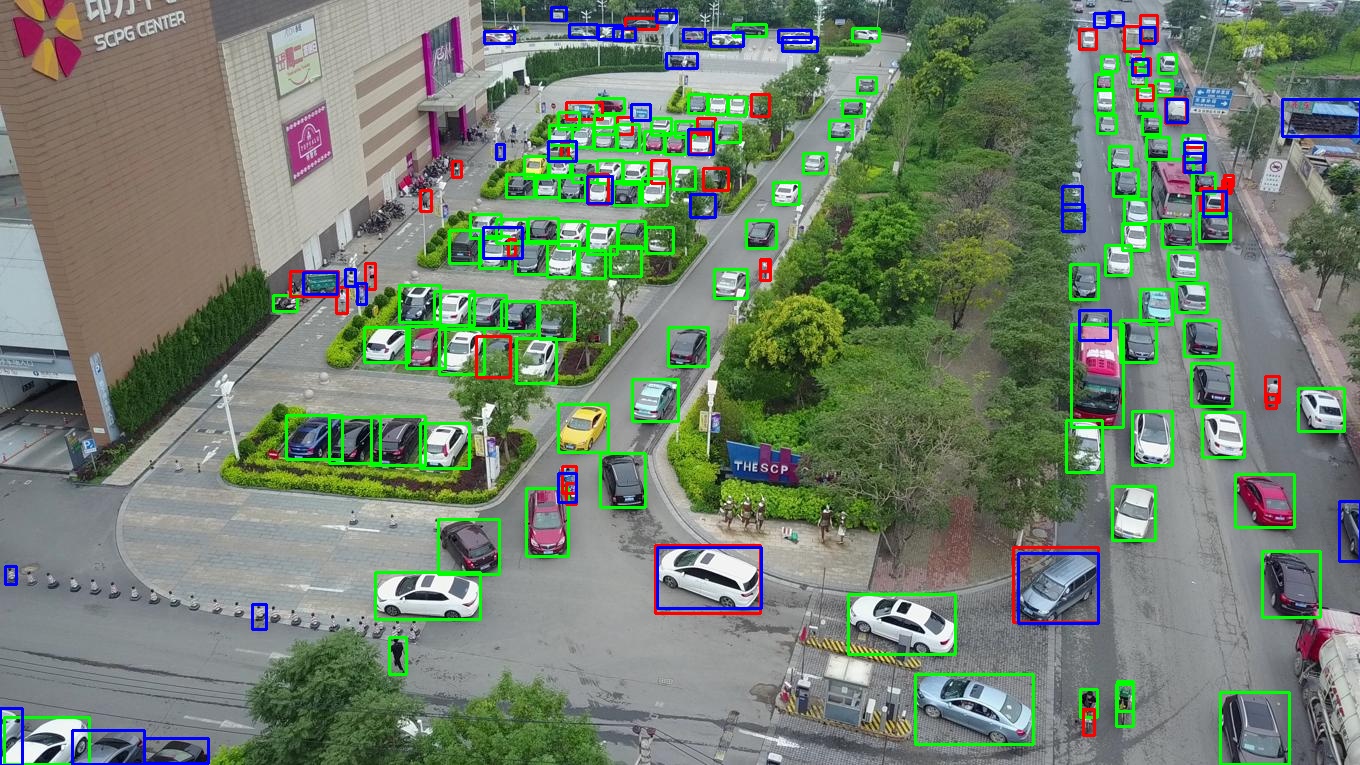}
        \vfill
        \scriptsize (b) YOLO11m
    \end{minipage}
    \begin{minipage}{0.24\textwidth}
        \centering
        \includegraphics[width=\linewidth]{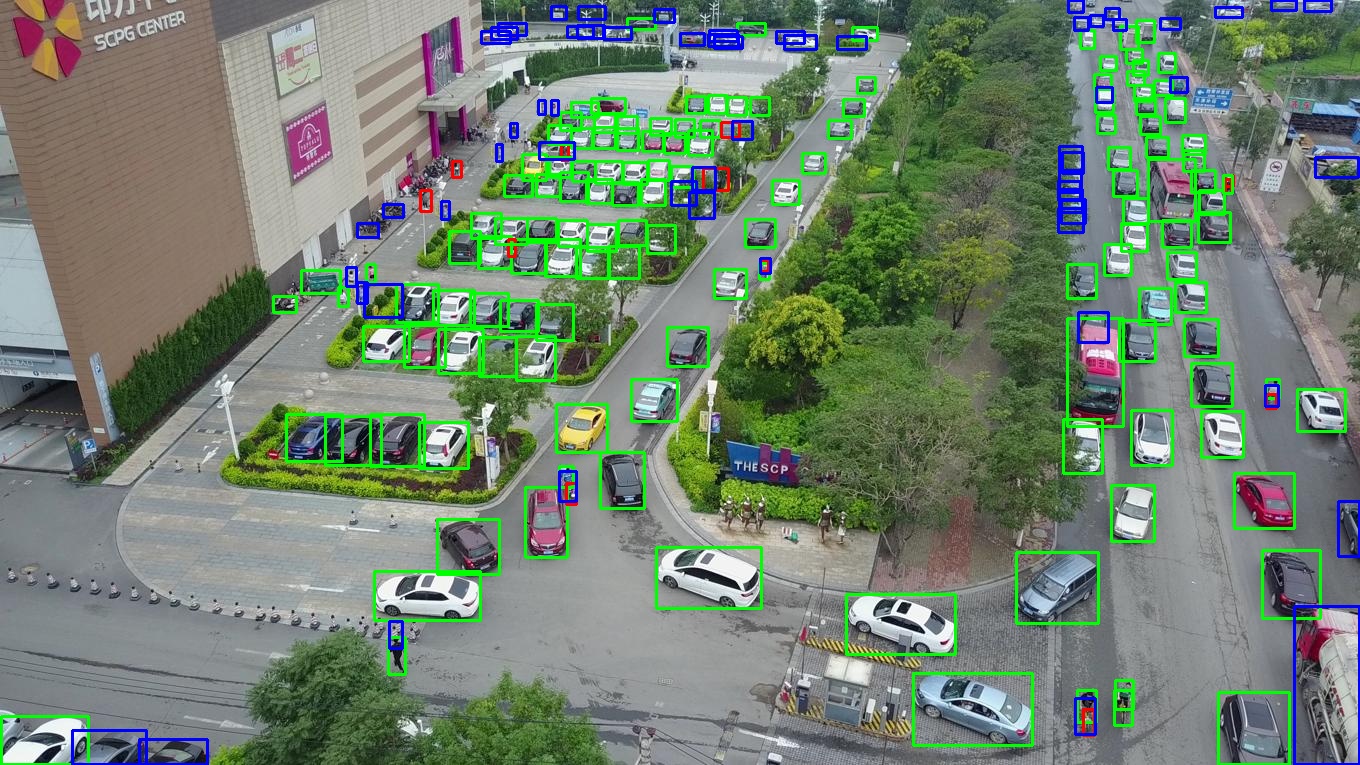}
        \vfill
        \scriptsize (c) D-Fine-M
    \end{minipage}
    \begin{minipage}{0.24\textwidth}
        \centering
        \includegraphics[width=\linewidth]{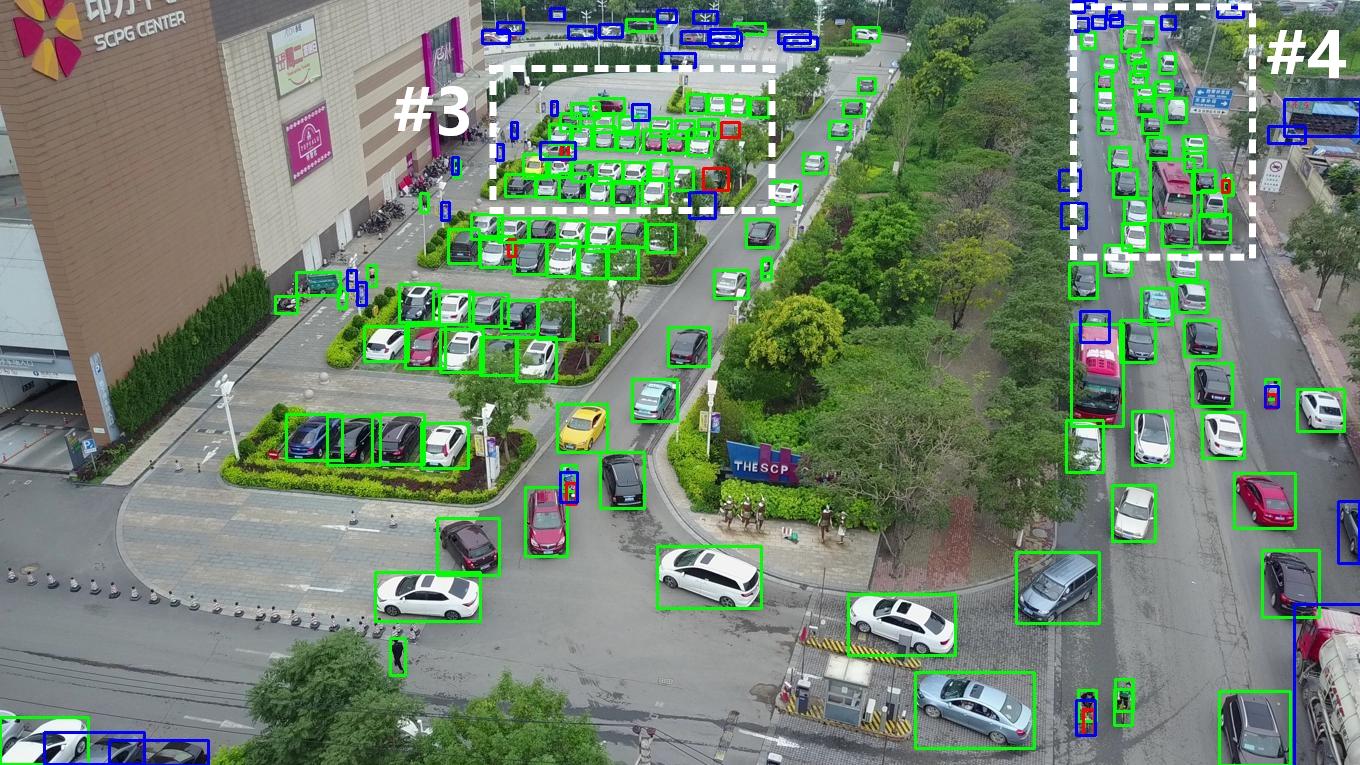}
        \vfill
        \scriptsize (d) Ours
    \end{minipage}

    \caption{Visual error analysis on the VisDrone 2019 dataset across two typical traffic scenes. FSDETR (Ours) effectively reduces blue FN boxes in high-density areas through enhanced frequency-spatial modeling. Legend: TP (Green), FP (Red), FN (Blue).}
    \label{fig:visdrone_qualitative}
\end{figure*}

\begin{figure*}[htbp]
    \centering
    \includegraphics[width=0.24\textwidth]{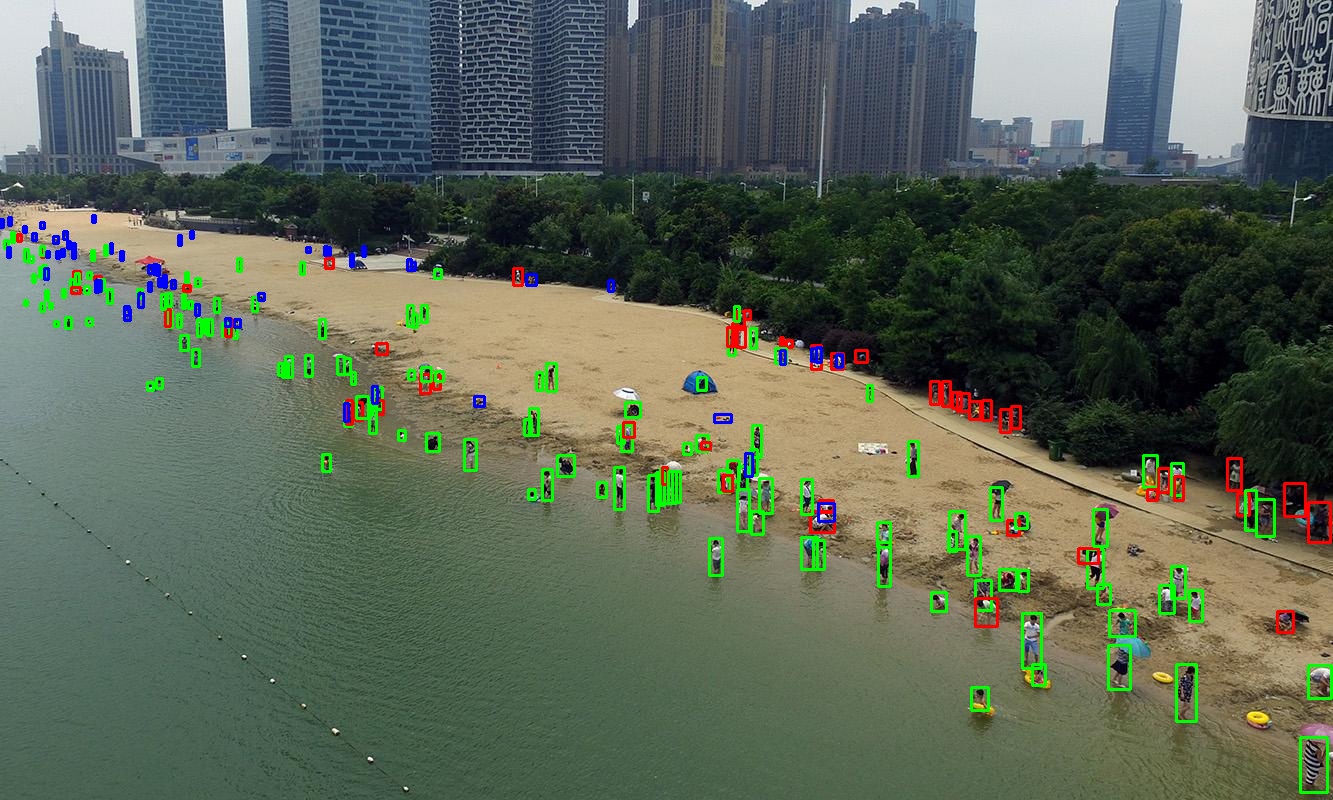}
    \includegraphics[width=0.24\textwidth]{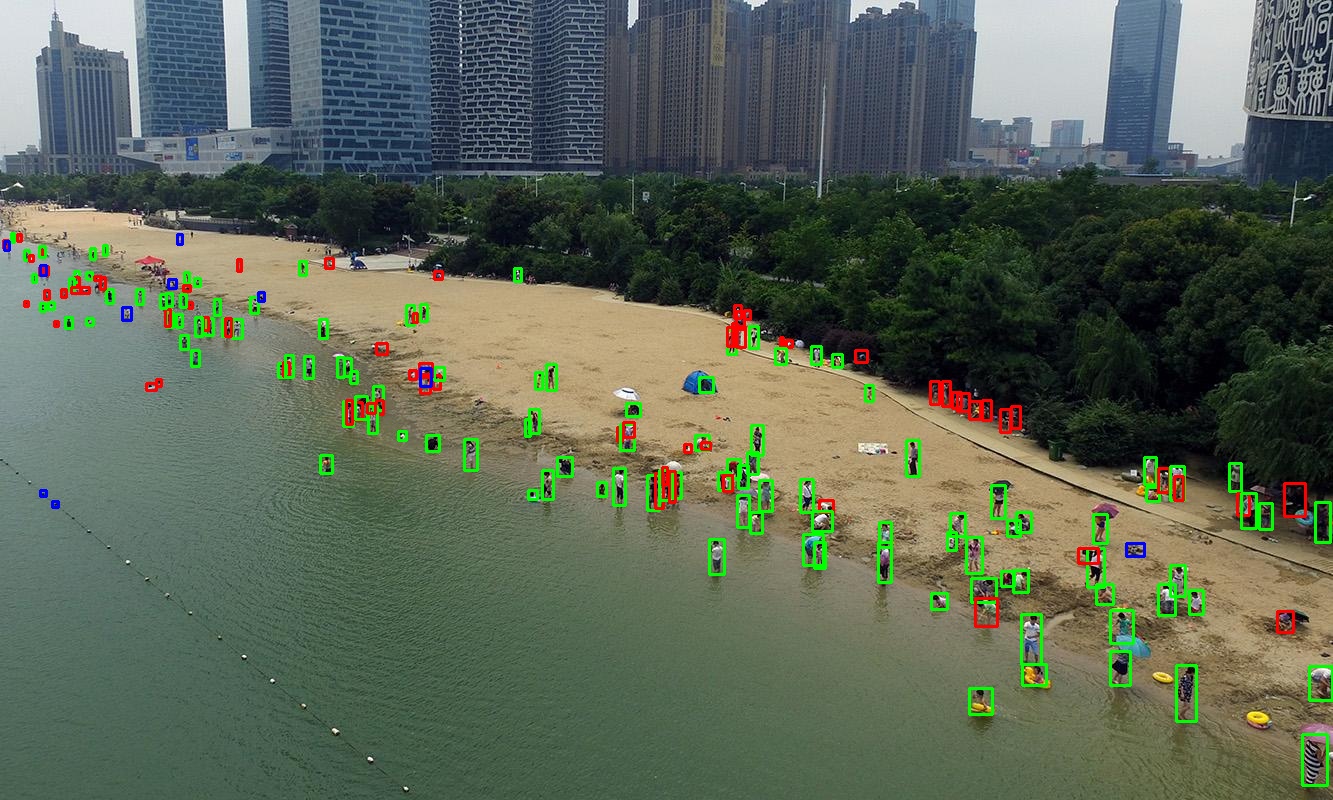}
    \includegraphics[width=0.24\textwidth]{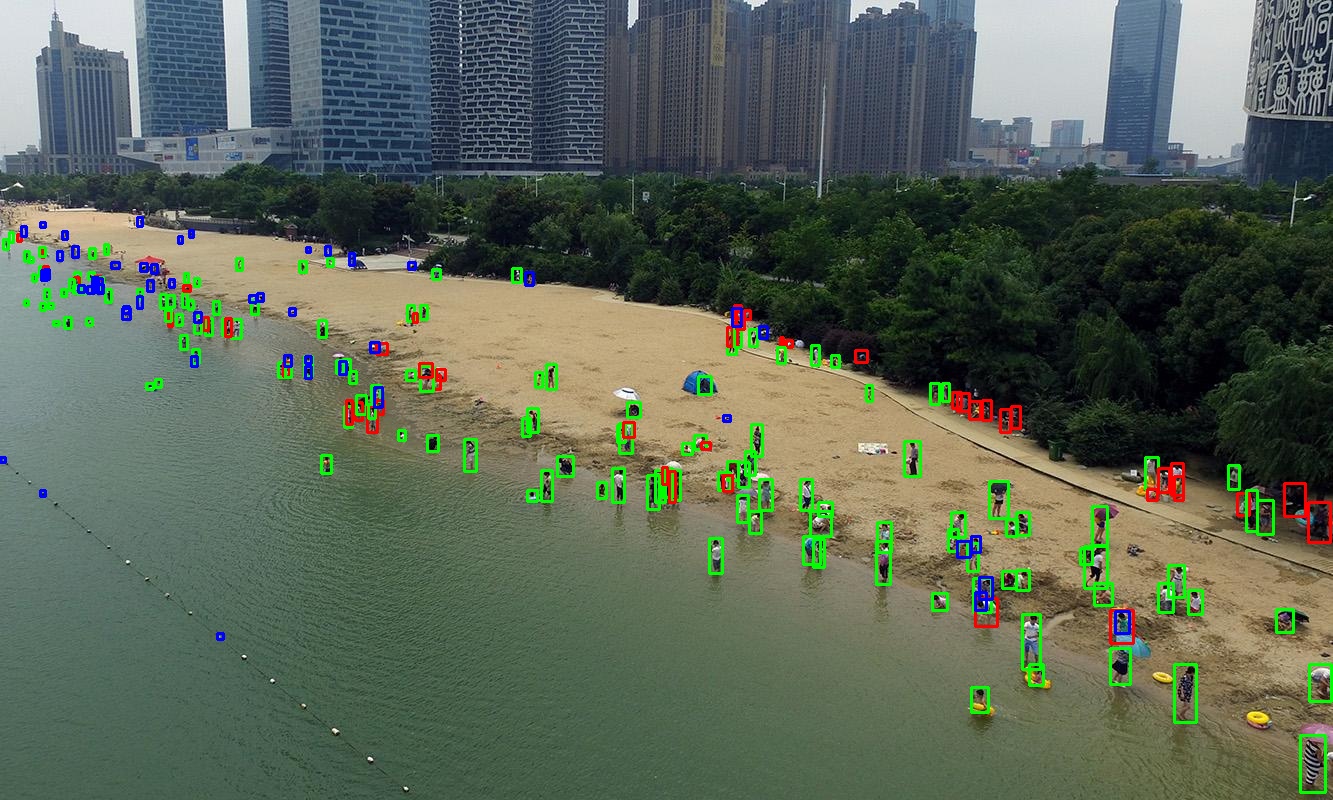}
    \includegraphics[width=0.24\textwidth]{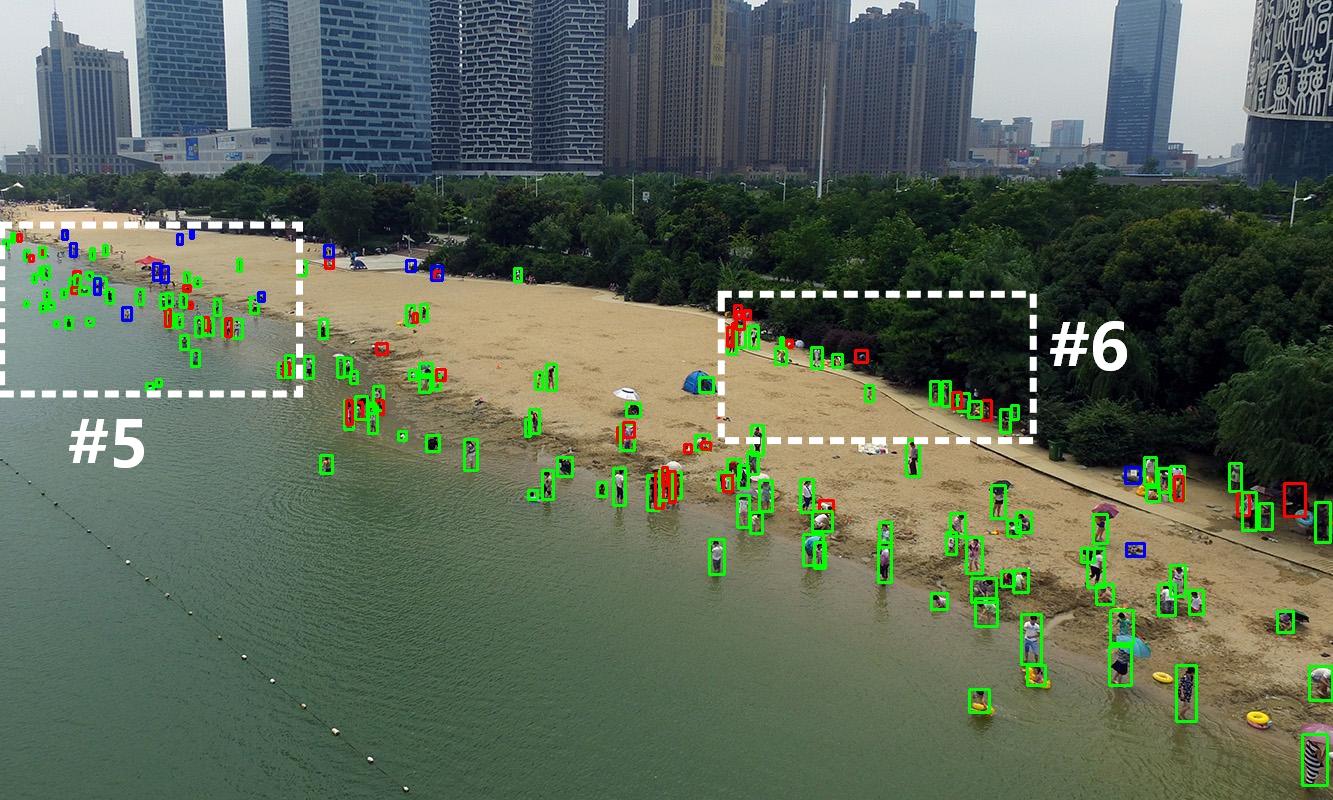}
    
    \vspace{2pt}
    
    \begin{minipage}{0.24\textwidth}
        \centering
        \includegraphics[width=\linewidth]{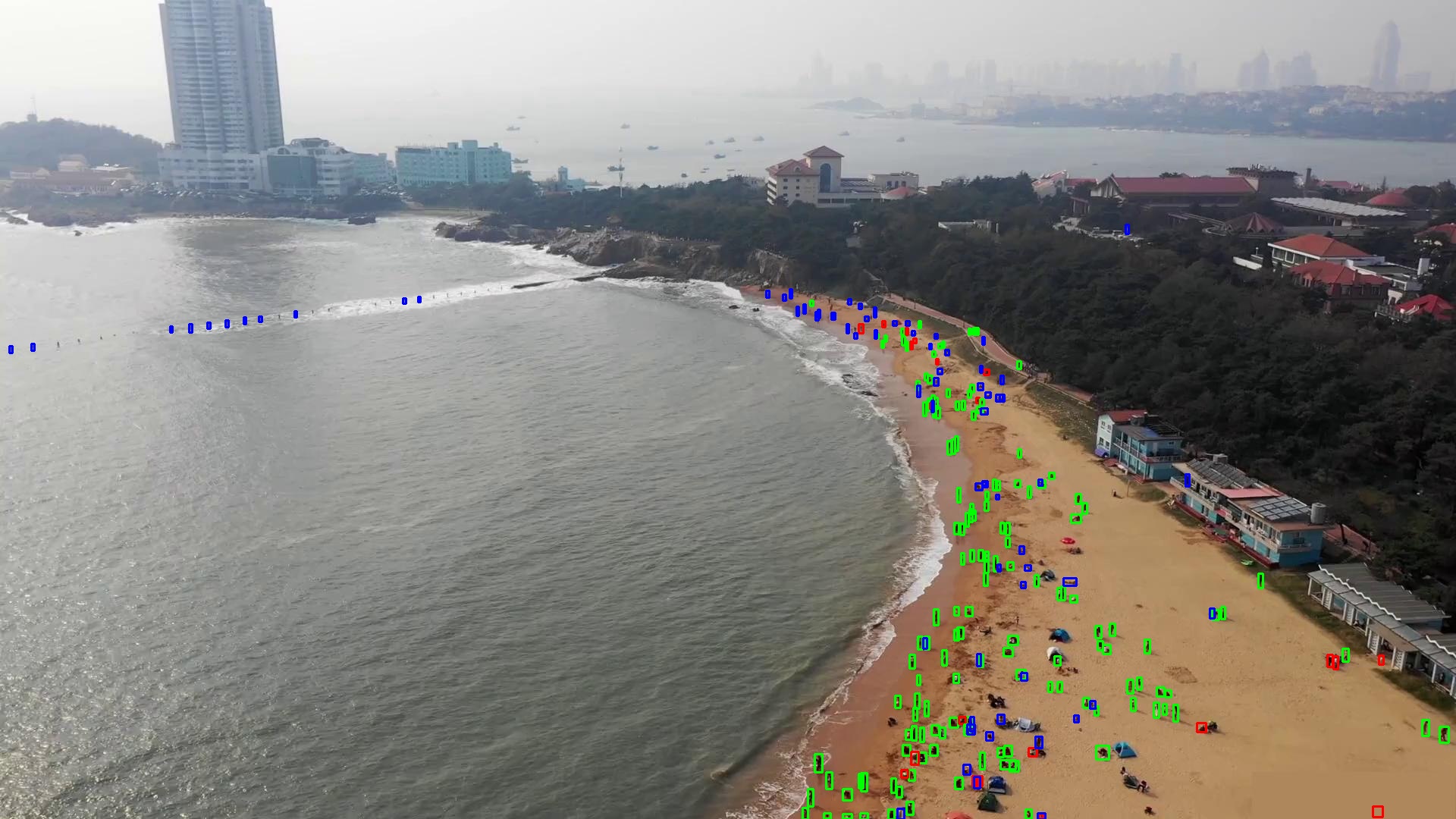}
        \vfill
        \scriptsize (a) Baseline
    \end{minipage}
    \begin{minipage}{0.24\textwidth}
        \centering
        \includegraphics[width=\linewidth]{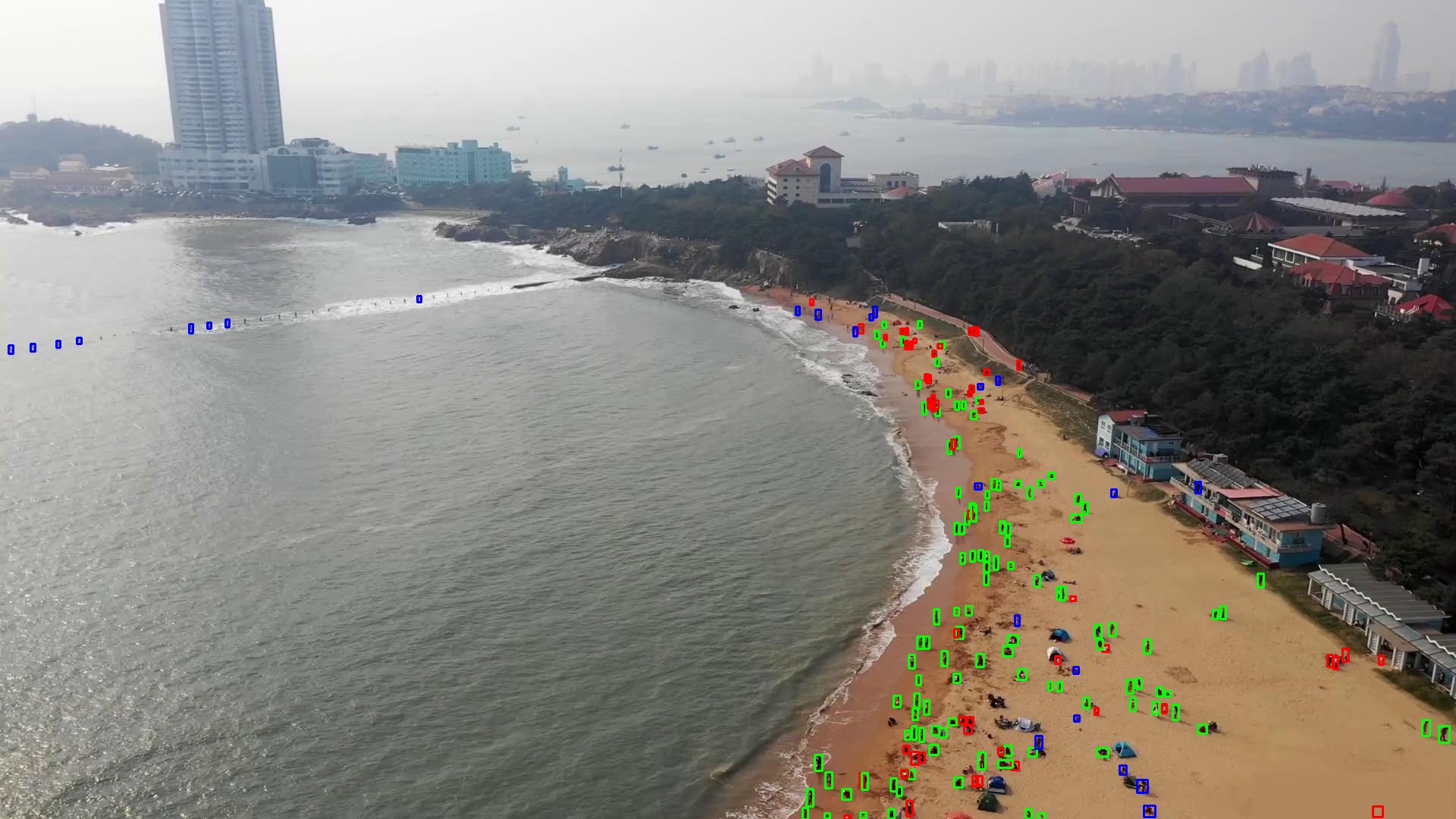}
        \vfill
        \scriptsize (b) YOLO11m
    \end{minipage}
    \begin{minipage}{0.24\textwidth}
        \centering
        \includegraphics[width=\linewidth]{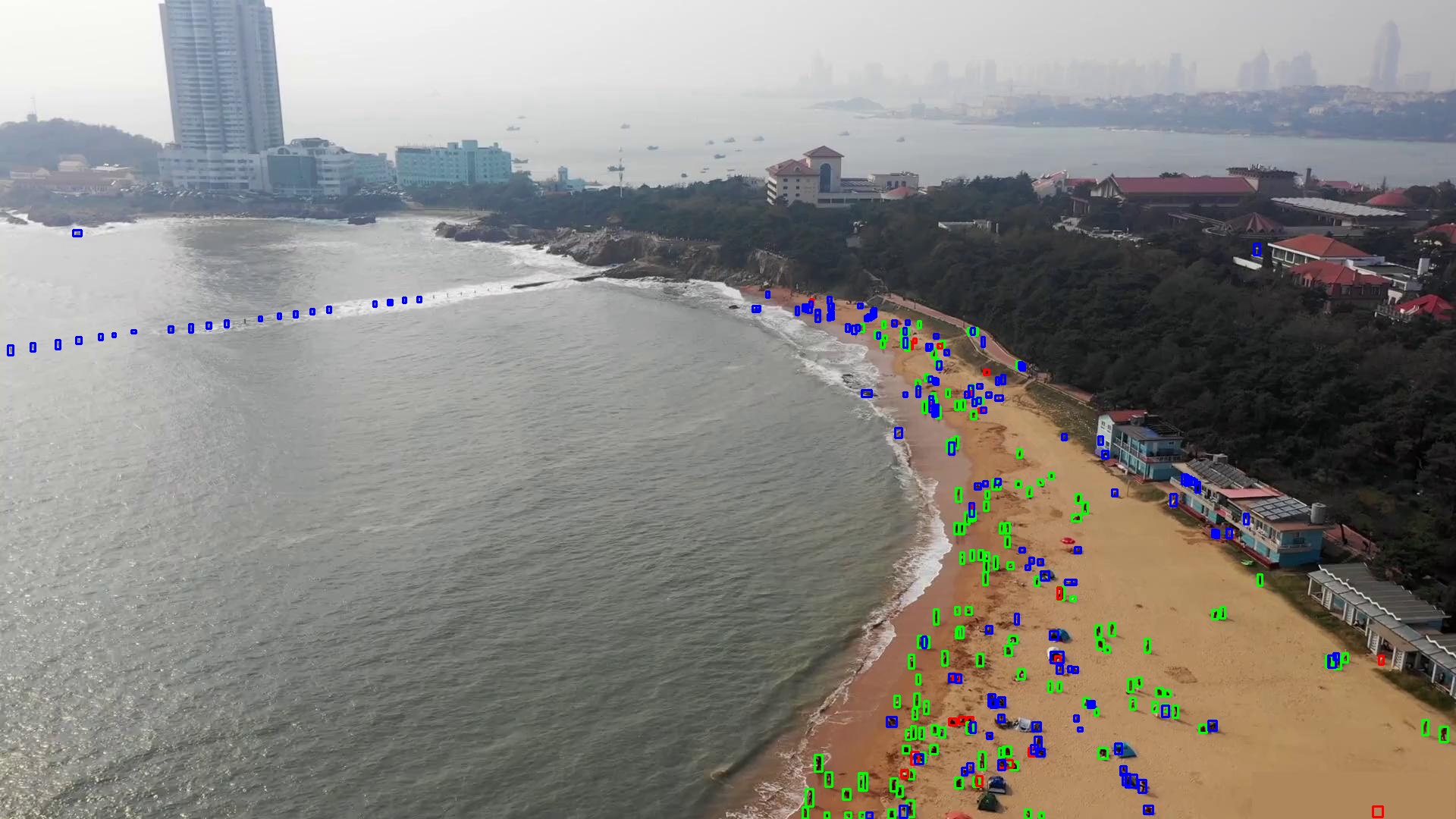}
        \vfill
        \scriptsize (c) D-Fine-M
    \end{minipage}
    \begin{minipage}{0.24\textwidth}
        \centering
        \includegraphics[width=\linewidth]{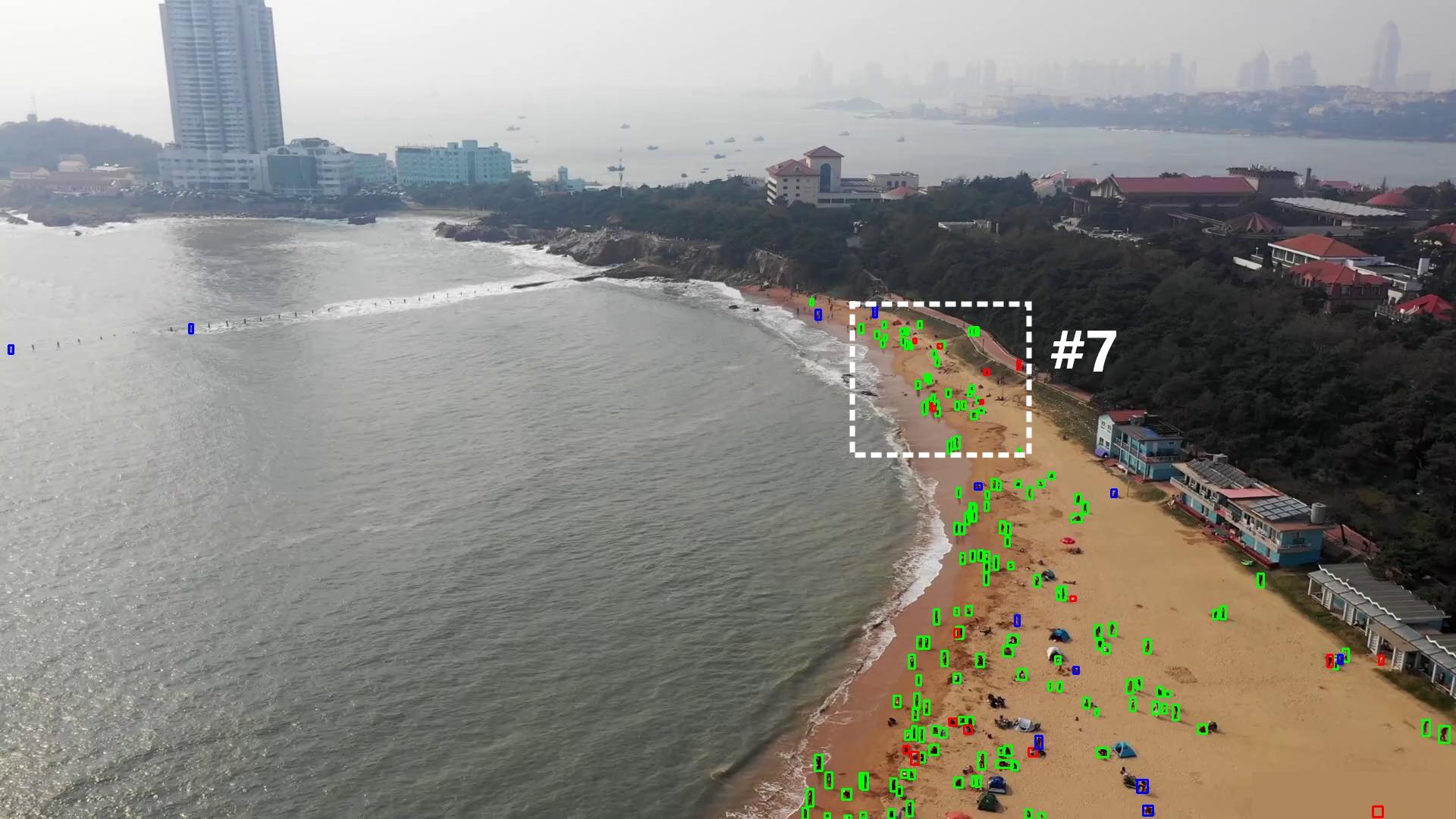}
        \vfill
        \scriptsize (d) Ours
    \end{minipage}

    \caption{Visual error analysis on the TinyPerson dataset. In these low-contrast seaside environments, FSDETR demonstrates superior target localization and lower false alarm rates (red boxes) for extremely small objects. Legend: TP (Green), FP (Red), FN (Blue).}
    \label{fig:tinyperson_qualitative}

    \vspace{-8pt}
\end{figure*}

\section{Experiments}

\subsection{Datasets and Implementation Details}

Experiments are conducted on two benchmark datasets with extreme scale variation: VisDrone 2019~\cite{Du_2019_ICCV} and TinyPerson~\cite{yu2020scale}. VisDrone 2019 contains UAV images with dense small objects under complex backgrounds, while TinyPerson focuses on long-range pedestrian detection with targets typically smaller than $20 \times 20$ pixels. 
FSDETR is implemented based on RT-DETR~\cite{zhao2023detrs} and trained on an NVIDIA GeForce RTX 4060 Ti GPU using AdamW with an initial learning rate of $1 \times 10^{-4}$, cosine annealing, 200 epochs, and a batch size of 4. Loss weights are set to $\lambda_{cls}=2$, $\lambda_{L1}=5$, and $\lambda_{IoU}=2$. For TinyPerson, the official image-slicing strategy is used during training and testing, and NMS is applied to merge sub-image predictions.

\subsection{Comparison with State-of-the-Art}

To evaluate FSDETR, comparisons are performed against a diverse set of state-of-the-art detectors, including representative CNN-based architectures, real-time YOLO variants, and Transformer-based models, alongside specialized algorithms optimized for small-scale targets. For a fair comparison, representative methods were reproduced under settings aligned as closely as possible with their official implementations and evaluation protocols.

\subsubsection{Quantitative comparison}
Experiments are evaluated using standard COCO metrics and the official TinyPerson~\cite{yu2020scale} evaluation protocols. As shown in Table~\ref{tab:visdrone}, FSDETR achieves a favorable balance between model complexity and detection accuracy. With only 14.7M parameters, representing a 26.5\% reduction from the RT-DETR-R18 baseline~\cite{zhao2023detrs}, it attains the best $AP_S$ of 13.9\% on VisDrone. On TinyPerson (Table~\ref{tab:tinyperson}), FSDETR achieves the best $AP_{50}^{tiny1}$ of 31.85\%, the best $AP_{50}^{tiny}$ of 48.95\%, and $AP_{25}^{tiny}$ of 72.80\%. These results indicate that the proposed frequency-spatial enhancement strategy is effective for improving small-object detection.

\subsubsection{Qualitative Comparison}
Qualitative evaluations on VisDrone 2019 and TinyPerson (Figs. \ref{fig:visdrone_qualitative}--\ref{fig:tinyperson_qualitative}) highlight FSDETR's robustness in challenging scenarios (Regions \#1--\#7). In VisDrone scenes (\#1--\#4), FSFPN effectively preserves weak signals often lost during downsampling, while SHAB maintains high detection density through long-range dependency modeling. Similarly, for sub-$20 \times 20$ pixel targets in TinyPerson's low-contrast environments (\#5--\#7), FSDETR mitigates feature blurring by integrating image slicing with high-frequency extraction. These observations align with the quantitative gains in $AP_S$ and $AP_{50}^{tiny1}$, visually validating its precision and robustness in tiny object detection.

\subsection{Ablation Study}
The efficacy of FSDETR's core components is validated through systematic ablation studies on the VisDrone 2019 dataset using RT-DETR-R18~\cite{zhao2023detrs} as the baseline, with results detailed in Table~\ref{tab:ablation}. Integrating SHAB yields a 1.2\% improvement in $AP_{50}$, demonstrating that lightweight self-attention effectively expands receptive fields and enhances the global context perception essential for distinguishing small targets. Subsequently, substituting the interaction mechanism with DA-AIFI~\cite{zhu2021deformable} adaptively focuses on sparse, information-rich regions via dynamic sampling, enhancing localization accuracy and geometric adaptability with negligible overhead.

\begin{table}[htbp]
\centering
\caption{Ablation experiments on component effectiveness}
\label{tab:ablation}
\resizebox{\columnwidth}{!}{%
\begin{tabular}{ccccccc}
\toprule
Exp. & SHAB & DA-AIFI & FSFPN (CFSB) & AP50 (\%) & AP (\%) & APS (\%) \\
\midrule
1 & & & & 36.3 & 20.8 & 11.3 \\
2 & \checkmark & & & 37.5 & 21.5 & 11.9 \\
3 & & \checkmark & & 37.1 & 21.2 & 11.6 \\
4 & & & \checkmark & 38.2 & 21.9 & 12.5 \\
5 & \checkmark & \checkmark & & 38.9 & 22.1 & 12.3 \\
6 & \textbf{\checkmark} & \textbf{\checkmark} & \textbf{\checkmark} & \textbf{40.5} & \textbf{22.7} & \textbf{13.9} \\
\bottomrule
\end{tabular}%
}
\end{table}

Notably, FSFPN with CFSB delivers the most substantial gain, increasing $AP_S$ and $AP_{50}$ by 1.2\% and 1.9\%, respectively. This confirms frequency-domain information complements spatial features, mitigating the loss of high-frequency textures during deep feature fusion~\cite{chi2020fast}. Ultimately, the complete architecture achieves a peak $AP_S$ of 13.9\%, highlighting that the deep coupling of spatial semantics and frequency details is crucial for recovering faint signals from small objects.

\section{CONCLUSION}

This study proposes FSDETR to address the challenge of small-object feature degradation stemming from excessive spatial compression in deep neural networks. The proposed architecture integrates SHAB to expand receptive fields and employs DA-AIFI to achieve precise geometric alignment, thereby establishing robust long-range dependencies with minimal computational overhead. Furthermore, FSFPN is introduced to leverage the deep coupling of spatial semantics and frequency-domain textures, explicitly capturing and amplifying weak signals typically overlooked by conventional architectures. Ablation studies and comparative analyses on the VisDrone 2019 and TinyPerson datasets demonstrate that FSDETR, with only 14.7M parameters, achieves competitive performance with clear gains on key small-object metrics. These experimental results validate the efficacy of the frequency-spatial collaborative strategy in enhancing detection precision and robustness for tiny targets in complex real-world environments.

\section*{Acknowledgment}
We acknowledge the use of Google Gemini~\cite{gemini} for text polishing and linguistic refinement. The generated content was verified by the authors to ensure accuracy and consistency.

\bibliographystyle{ieeetr}  
\bibliography{references}   

@inproceedings{lin2017feature,
  author    = {T.-Y. Lin and P. Doll{\'a}r and R. Girshick and K. He and B. Hariharan and S. Belongie},
  title     = {Feature pyramid networks for object detection},
  booktitle = {Proc. IEEE Conf. Comput. Vis. Pattern Recognit. ({CVPR})},
  year      = {2017},
  pages     = {2117--2125}
}

@inproceedings{liu2018path,
  author    = {S. Liu and L. Qi and H. Qin and J. Shi and J. Jia},
  title     = {Path aggregation network for instance segmentation},
  booktitle = {Proc. IEEE Conf. Comput. Vis. Pattern Recognit. ({CVPR})},
  year      = {2018},
  pages     = {8759--8768}
}

@inproceedings{zhu2021deformable,
  author    = {X. Zhu and W. Su and L. Lu and B. Li and X. Wang and J. Dai},
  title     = {Deformable {DETR}: Deformable transformers for end-to-end object detection},
  booktitle = {Proc. Int. Conf. Learn. Represent. ({ICLR})},
  year      = {2021}
}

@inproceedings{zhao2023detrs,
  author    = {H. Zhao and J. Zhang and Y. Zhao and P. Li and C.-C. Loy and D. Lin and J. Jia},
  title     = {{DETRs} beat {YOLOs} on real-time object detection},
  booktitle = {Proc. IEEE Conf. Comput. Vis. Pattern Recognit. ({CVPR})},
  year      = {2023},
  pages     = {17063--17073}
}

@inproceedings{chi2020fast,
  author    = {L. Chi and B. Jiang and Y. Mu},
  title     = {Fast fourier convolution},
  booktitle = {Adv. Neural Inf. Process. Syst. ({NeurIPS})},
  volume    = {33},
  year      = {2020},
  pages     = {4479--4490}
}

@InProceedings{Zhu_2021_ICCV,
    author    = {Zhu, Xingkui and Lyu, Shuchang and Wang, Xu and Zhao, Qi},
    title     = {TPH-YOLOv5: Improved YOLOv5 Based on Transformer Prediction Head for Object Detection on Drone-Captured Scenarios},
    booktitle = {Proceedings of the IEEE/CVF International Conference on Computer Vision (ICCV) Workshops},
    month     = {October},
    year      = {2021},
    pages     = {2778-2788}
}

@inproceedings{wang2020cspnet,
  title={CSPNet: A new backbone that can enhance learning capability of CNN},
  author={Wang, Chien-Yao and Liao, Hong-Yuan Mark and Wu, Yueh-Hua and Chen, Ping-Yang and Hsieh, Jun-Wei and Yeh, I-Hau},
  booktitle={Proceedings of the IEEE/CVF conference on computer vision and pattern recognition workshops},
  pages={390--391},
  year={2020}
}

@article{wang2024yolov10,
  title={Yolov10: Real-time end-to-end object detection},
  author={Wang, Ao and Chen, Hui and Liu, Lihao and Chen, Kai and Lin, Zijia and Han, Jungong and others},
  journal={Advances in Neural Information Processing Systems},
  volume={37},
  pages={107984--108011},
  year={2024}
}

@inproceedings{spd-conv2022,
  title={No More Strided Convolutions or Pooling: A New {CNN} Building Block for Low-Resolution Images and Small Objects},
  author={Raja Sunkara and Tie Luo},
  booktitle={European Conference on Machine Learning and Principles and Practice of Knowledge Discovery in Databases (ECML PKDD)},
  month=Sep,
  year={2022},
  pages={443-459}
}

@article{zhang2022dino,
  title={Dino: Detr with improved denoising anchor boxes for end-to-end object detection},
  author={Zhang, Hao and Li, Feng and Liu, Shilong and Zhang, Lei and Su, Hang and Zhu, Jun and Ni, Lionel M and Shum, Heung-Yeung},
  journal={arXiv preprint arXiv:2203.03605},
  year={2022}
}

@inproceedings{wang2021pyramid,
  title={Pyramid vision transformer: A versatile backbone for dense prediction without convolutions},
  author={Wang, Wenhai and Xie, Enze and Li, Xiang and Fan, Deng-Ping and Song, Kaitao and Liang, Ding and Lu, Tong and Luo, Ping and Shao, Ling},
  booktitle={Proceedings of the IEEE/CVF international conference on computer vision},
  pages={568--578},
  year={2021}
}

@misc{Jocher_Ultralytics_YOLO_2023,
author = {Jocher, Glenn and Qiu, Jing and Chaurasia, Ayush},
license = {AGPL-3.0},
month = jan,
title = {{Ultralytics YOLO}},
url = {https://github.com/ultralytics/ultralytics},
version = {8.0.0},
year = {2023}
}

@article{chen2023run,
  title={Run, Don't Walk: Chasing Higher FLOPS for Faster Neural Networks},
  author={Chen, Jierun and Kao, Shiu-hong and He, Hao and Zhuo, Weipeng and Wen, Song and Lee, Chul-Ho and Chan, S-H Gary},
  journal={arXiv preprint arXiv:2303.03667},
  year={2023}
}

@article{rao2021global,
  title={Global filter networks for image classification},
  author={Rao, Yongming and Zhao, Wenliang and Zhu, Zheng and Lu, Jiwen and Zhou, Jie},
  journal={Advances in neural information processing systems},
  volume={34},
  pages={980--993},
  year={2021}
}

@article{Lin2017FocalLF,
  title={Focal Loss for Dense Object Detection},
  author={Tsung-Yi Lin and Priya Goyal and Ross B. Girshick and Kaiming He and Piotr Doll{\'a}r},
  journal={IEEE Transactions on Pattern Analysis and Machine Intelligence},
  year={2017},
  volume={42},
  pages={318-327},
  url={https://api.semanticscholar.org/CorpusID:206771220}
}

@inproceedings{cai2018cascade,
  title={Cascade r-cnn: Delving into high quality object detection},
  author={Cai, Zhaowei and Vasconcelos, Nuno},
  booktitle={Proceedings of the IEEE conference on computer vision and pattern recognition},
  pages={6154--6162},
  year={2018}
}

@misc{yolo11_ultralytics,
  author = {Glenn Jocher and Jing Qiu},
  title = {Ultralytics YOLO11},
  version = {11.0.0},
  year = {2024},
  url = {https://github.com/ultralytics/ultralytics},
  orcid = {0000-0001-5950-6979, 0000-0003-3783-7069},
  license = {AGPL-3.0}
}

@article{tian2025yolo12,
  title={YOLO12: Attention-Centric Real-Time Object Detectors},
  author={Tian, Yunjie and Ye, Qixiang and Doermann, David},
  journal={arXiv preprint arXiv:2502.12524},
  year={2025}
}

@misc{peng2024dfine,
      title={D-FINE: Redefine Regression Task in DETRs as Fine-grained Distribution Refinement},
      author={Yansong Peng and Hebei Li and Peixi Wu and Yueyi Zhang and Xiaoyan Sun and Feng Wu},
      year={2024},
      eprint={2410.13842},
      archivePrefix={arXiv},
      primaryClass={cs.CV}
}

@misc{lv2024rtdetrv2improvedbaselinebagoffreebies,
      title={RT-DETRv2: Improved Baseline with Bag-of-Freebies for Real-Time Detection Transformer}, 
      author={Wenyu Lv and Yian Zhao and Qinyao Chang and Kui Huang and Guanzhong Wang and Yi Liu},
      year={2024},
      eprint={2407.17140},
      archivePrefix={arXiv},
      primaryClass={cs.CV},
      url={https://arxiv.org/abs/2407.17140}, 
}

@inproceedings{xu2022rfla,
  title={RFLA: Gaussian receptive field based label assignment for tiny object detection},
  author={Xu, Chang and Wang, Jinwang and Yang, Wen and Yu, Huai and Yu, Lei and Xia, Gui-Song},
  booktitle={European Conference on Computer Vision},
  pages={526--543},
  year={2022},
  organization={Springer}
}

@inproceedings{zhang2021varifocalnet,
  title={Varifocalnet: An iou-aware dense object detector},
  author={Zhang, Haoyang and Wang, Ying and Dayoub, Feras and Sunderhauf, Niko},
  booktitle={Proceedings of the IEEE/CVF conference on computer vision and pattern recognition},
  pages={8514--8523},
  year={2021}
}

@article{zhang2022focal,
  title={Focal and efficient IOU loss for accurate bounding box regression},
  author={Zhang, Yi-Fan and Ren, Weiqiang and Zhang, Zhang and Jia, Zhen and Wang, Liang and Tan, Tieniu},
  journal={Neurocomputing},
  volume={506},
  pages={146--157},
  year={2022},
  publisher={Elsevier}
}

@InProceedings{Du_2019_ICCV,
author = {Du, Dawei and Zhu, Pengfei and Wen, Longyin and Bian, Xiao and Lin, Haibin and Hu, Qinghua and Peng, Tao and Zheng, Jiayu and Wang, Xinyao and Zhang, Yue and Bo, Liefeng and Shi, Hailin and Zhu, Rui and Kumar, Aashish and Li, Aijin and Zinollayev, Almaz and Askergaliyev, Anuar and Schumann, Arne and Mao, Binjie and Lee, Byeongwon and Liu, Chang and Chen, Changrui and Pan, Chunhong and Huo, Chunlei and Yu, Da and Cong, DeChun and Zeng, Dening and Reddy Pailla, Dheeraj and Li, Di and Wang, Dong and Cho, Donghyeon and Zhang, Dongyu and Bai, Furui and Jose, George and Gao, Guangyu and Liu, Guizhong and Xiong, Haitao and Qi, Hao and Wang, Haoran and Qiu, Heqian and Li, HongLiang and Lu, Huchuan and Kim, Ildoo and Kim, Jaekyum and Shen, Jane and Lee, Jihoon and Ge, Jing and Xu, Jingjing and Zhou, Jingkai and Meier, Jonas and Won Choi, Jun and Hu, Junhao and Zhang, Junyi and Huang, Junying and Huang, Kaiqi and Wang, Keyang and Sommer, Lars and Jin, Lei and Zhang, Lei},
title = {VisDrone-DET2019: The Vision Meets Drone Object Detection in Image Challenge Results},
booktitle = {Proceedings of the IEEE/CVF International Conference on Computer Vision (ICCV) Workshops},
month = {Oct},
year = {2019}
}

@inproceedings{yu2020scale,
  title={Scale match for tiny person detection},
  author={Yu, Xuehui and Gong, Yuqi and Jiang, Nan and Ye, Qixiang and Han, Zhenjun},
  booktitle={Proceedings of the IEEE/CVF winter conference on applications of computer vision},
  pages={1257--1265},
  year={2020}
}

@article{2019FreeAnchor,
  title={FreeAnchor: Learning to Match Anchors for Visual Object Detection},
  author={ Zhang, Xiaosong  and  Wan, Fang  and  Liu, Chang  and  Ji, Rongrong  and  Ye, Qixiang },
  journal={arXiv},
  year={2019},
}

@INPROCEEDINGS{Wang2025CCD,
  author={Wang, Dong and Jiang, Yuntian and Wang, Shixinyu},
  booktitle={2025 International Joint Conference on Neural Networks (IJCNN)}, 
  title={The Improved Algorithm of RT-DETR Based on Multi-Head Attention Mechanism}, 
  year={2025},
  volume={},
  number={},
  pages={1-8},
  doi={10.1109/IJCNN64981.2025.11228131}}

@INPROCEEDINGS{Guo2025FDSI,
  author={Guo, Tianyu and Song, Qingzeng and Xue, Yongjiang and Qiao, Fei},
  booktitle={2025 International Joint Conference on Neural Networks (IJCNN)}, 
  title={FDSI-RTDETR : A Lightweight Unmanned Aerial Vehicle (UAV) Aerial Image Small Object Detection Network}, 
  year={2025},
  volume={},
  number={},
  pages={1-8},
  doi={10.1109/IJCNN64981.2025.11228989}}

@INPROCEEDINGS{Liu2025GCGP,
  author={Liu, Jinsheng and Yan, Tao and Chen, Xianglong and Zhang, Zhineng and Wang, Chenglong and Huang, Weilong},
  booktitle={2025 International Joint Conference on Neural Networks (IJCNN)}, 
  title={GCGP-YOLO: Global-Local and Channel Grouping Perception Network for Small Object Detection based on YOLOv8}, 
  year={2025},
  volume={},
  number={},
  pages={1-10},
  doi={10.1109/IJCNN64981.2025.11227737}}

@inproceedings{li2024rethinking,
  title={Rethinking Features-Fused-Pyramid-Neck for Object Detection},
  author={Li, Hulin},
  booktitle={European Conference on Computer Vision},
  pages={74--90},
  year={2024},
  organization={Springer}
}

@misc{gemini,
  author = {Google},
  title = {Gemini},
  year = {2026},
  howpublished = {[Large language model]. Available: \url{https://gemini.google.com}},
  note = {Accessed: Feb. 1, 2026}
}

\end{document}